\documentclass[11pt]{article}

\usepackage[final]{acl}

\usepackage{times}
\usepackage{latexsym}

\usepackage[T1]{fontenc}

\usepackage[utf8]{inputenc}

\NewDocumentCommand{\heng}
{ mO{} }{\textcolor{red}{\textsuperscript{\textit{Heng}}\textsf{\textbf{\small[#1]}}}}

\usepackage{microtype}

\usepackage{inconsolata}

\usepackage{graphicx}
\usepackage{booktabs}
\usepackage{amsmath}
\usepackage{xcolor}
\usepackage[table]{xcolor}
\usepackage{multirow}
\usepackage{amsfonts}
\usepackage{array}
\usepackage{tabularx}
\usepackage{multirow}
\usepackage{booktabs}
\usepackage{makecell}
\usepackage{listings}
\usepackage{tcolorbox}
\usepackage{enumitem}
\usepackage{float}

\definecolor{lightgrey}{RGB}{251, 202, 169}

\title{Learning to Predict Future-Aligned Research Proposals \\ with Language Models}

\author{
Heng Wang, Pengcheng Jiang, Jiashuo Sun, \textbf{{Zhiyi Shi}}, \\ \textbf{{Haofei Yu}}, \textbf{{Jiawei Han}}, \textbf{{Heng Ji}}  \\
University of Illinois Urbana-Champaign\\
\texttt{\{heng6, hanj, hengji\}@illinois.edu} \\
}

\begin{document}
\maketitle
\begin{abstract}
Large language models (LLMs) are increasingly used to assist ideation in research, but evaluating the quality of LLM-generated research proposals remains difficult: novelty and soundness are hard to measure automatically, and large-scale human evaluation is costly. 
We propose a verifiable alternative by reframing proposal generation as a time-sliced scientific forecasting problem.
Given a research question and inspiring papers available before a cutoff time, the model generates a structured proposal and is evaluated by whether it anticipates research directions that appear in papers published after the time.
We operationalize this objective with the Future Alignment Score (FAS), computed via retrieval and LLM-based semantic scoring against a held-out future corpus.
To train models, we build a time-consistent dataset of 21{,}835 paper occurrences across 3{,}642 instances from targets and their pre-cutoff citations, and synthesize reasoning traces that teach gap identification and inspiration borrowing.
Across Llama-3.1 and Qwen2.5 models, future-aligned tuning improves future alignment over unaligned baselines (up to +10.6\% overall FAS), and domain-expert human evaluation corroborates improved proposal quality.
Finally, we demonstrate practical impact by implementing two model-generated proposals with a code agent, obtaining 4.17\% accuracy gain on MATH from a new prompting strategy and consistent improvements for a novel model-merging method. 
Our code and data are publicly available at \url{https://github.com/Arthur-Heng/future-aligned-proposals}.
\end{abstract}

\begin{figure}[t]
    \centering
    \includegraphics[width=0.95\linewidth]{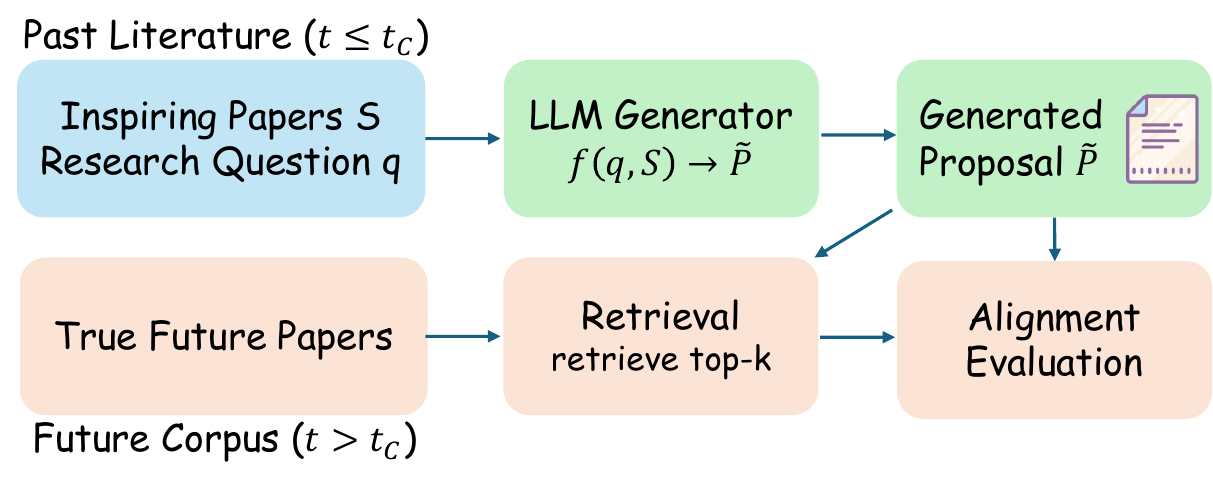}
    \vspace{-2mm}
    \caption{Given inspiring papers $S$ and a research question $q$ available before a cutoff time $t_C$, the model generates a proposal $\tilde{P}$. We evaluate whether the proposal anticipates future human research directions by comparing it against papers published after $t_C$ using retrieval and LLM-based semantic alignment.
    }
    \label{fig:teaser}
    
    \vspace{-3mm}
\end{figure}

\vspace{-1mm}
\section{Introduction}

Large language models (LLMs) are increasingly used to assist researchers in idea exploration, literature synthesis, and research proposal drafting~\citep{wang-etal-2024-scimon,si2025can}. 
Recent systems can identify potential research gaps~\citep{liu2026sci}, estimate or generate experimental outcomes~\citep{wen2025predicting}, and even produce end-to-end papers and code~\citep{lu2024ai,yu2025researchtown,yu-etal-2025-tinyscientist}, raising the possibility that LLMs could meaningfully accelerate scientific discovery. 
However, it remains unclear how to train and compare such systems at scale because there is no reliable automated objective for assessing the quality of generated research proposals.

This difficulty stems from the nature of proposal generation itself. Unlike question answering or summarization, research proposal generation has no single ground-truth output. 
Key attributes of a strong proposal---such as novelty, technical soundness, feasibility, and long-term impact---are subjective and often only verifiable retrospectively. 
As a result, automatic metrics, including LLM-based judges, are only weakly aligned with these properties, while human evaluation is costly, inconsistent, and difficult to scale~\citep{si2026the}. 
Consequently, it remains unclear what learning objective should guide models toward generating genuinely useful proposals rather than merely fluent and internally consistent text.

To address this challenge, we reframe proposal generation as a \emph{time-sliced scientific forecasting} problem (Figure~\ref{fig:teaser}). Instead of directly optimizing subjective notions such as novelty or creativity, we ask a verifiable question: \emph{can a model generate proposals that anticipate research directions later pursued by human researchers?} Concretely, given a research question $q$ and inspiring papers $S$ available before a cutoff time $t_C$, the model generates a structured proposal and is evaluated by its semantic alignment with papers published after $t_C$. If a proposal aligns strongly with future publications, it suggests that the model has captured research trajectories that later proved scientifically meaningful, rather than merely producing arbitrary creative text. We therefore treat future alignment as a scalable surrogate for proposal quality.

We operationalize this idea with the \textbf{Future Alignment Score (FAS)}, defined as the maximum semantic alignment between a generated proposal and top retrieved papers from a held-out future corpus, computed via retrieval and LLM-based semantic scoring. By turning proposal generation into a forecasting task in semantic space, this formulation provides a scalable automatic evaluation signal grounded in publication timelines.

To enable learning under this objective, we construct large-scale \emph{time-consistent supervision} from published papers. For each target paper, we extract a leakage-controlled research question and a set of inspiring citations, then synthesize a structured proposal that simulates a plausible pre-publication research plan. Beyond final proposal text, the model must also learn how to transform prior work into a new research direction by identifying gaps in the literature and repurposing useful ideas, which is a necessary process in human research. To capture this process, we synthesize \emph{citation-grounded reasoning traces} and introduce stepwise reasoning supervision that decomposes proposal generation into stages of problem formulation, method design, and experimental planning.

Experiments show that future-aligned supervision consistently improves future alignment over baselines, and ablation studies demonstrate that stepwise reasoning supervision yields additional gains. We further validate this framework beyond FAS: domain-expert human evaluation shows that Stepwise CoT proposals are preferred over prompting baselines across soundness, excitement, and overall assessment, while remaining competitive with human-derived proposals. Finally, to test whether generated proposals translate into executable research, we implement two proposals produced by our best model in an LLM-assisted coding environment, obtaining measurable improvements compared to baselines.

Our contributions can be summarized as:
\begin{itemize}[leftmargin=*]
    \item We introduce a time-sliced formulation of research proposal generation that evaluates proposals by their semantic match to future published papers, providing a scalable automatic signal grounded in future publications.

    \item We construct time-consistent supervision from published papers and their pre-cutoff citations, producing a dataset of 21{,}835 paper occurrences across 3{,}642 instances and synthesizing structured proposal targets and reasoning traces that capture gap analysis and inspiration borrowing.
    \item Across multiple LLM families, supervised fine-tuning substantially improves future alignment, and domain-expert human evaluation corroborates improvements in perceived proposal quality; we further demonstrate practical impact by implementing two model-generated proposals.
\end{itemize}

\section{Method}
\label{sec:method}
\begin{figure*}[t]
    \centering
    \includegraphics[width=0.9\linewidth]{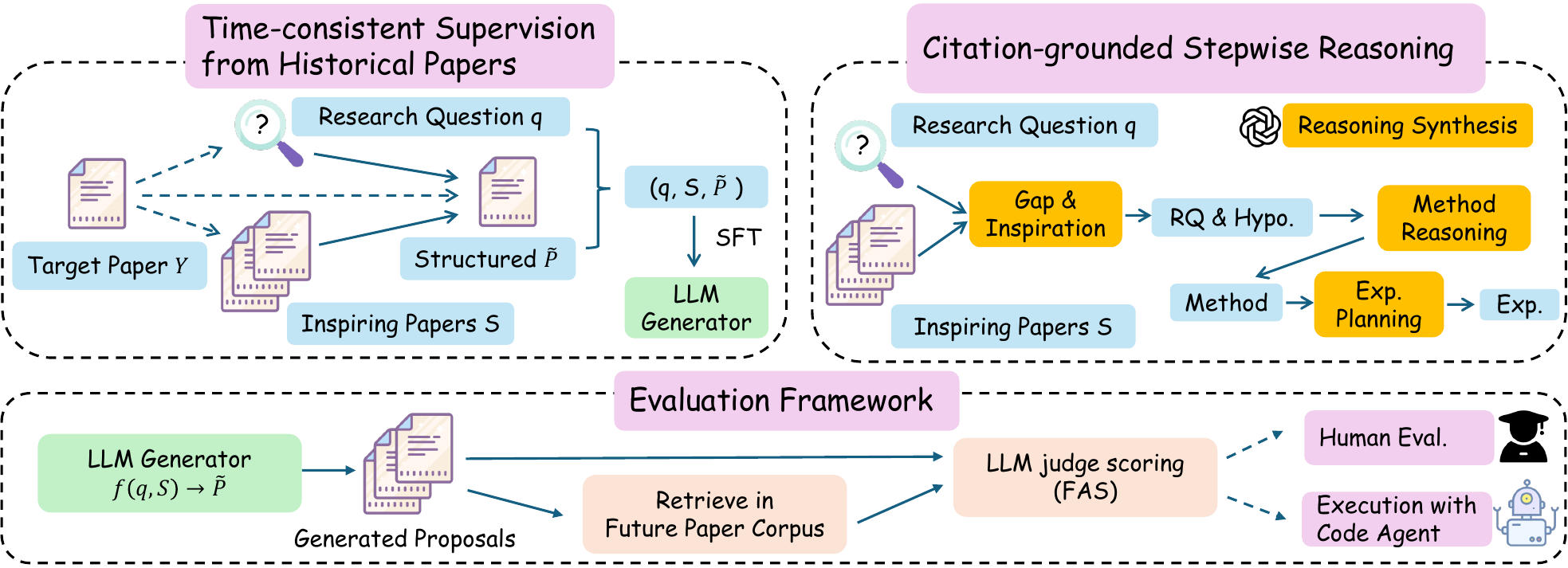}
    \vspace{-2mm}
    \caption{Overview of the proposed future-aligned learning framework. \emph{Time-consistent supervision} constructs training data from historical papers without future leakage, and \emph{citation-grounded stepwise reasoning} decomposes proposal generation into staged scientific planning. Together, these enable LoRA-based supervised fine-tuning of a proposal generator, which is evaluated by Future Alignment Score (FAS) against a held-out future corpus and further validated through human evaluation and execution-based case studies.}

    \label{fig:overview}
    \vspace{-4mm}
\end{figure*}

Research proposal generation is difficult to evaluate and supervise at scale. We address this with a future-aligned learning framework (Figure~\ref{fig:overview}) built on three key ideas: (1) \emph{future alignment} as a verifiable surrogate for proposal quality, (2) \emph{time-consistent supervision} synthesized from historical papers without future leakage, and (3) \emph{citation-grounded stepwise reasoning} that decomposes proposal generation into staged scientific planning.

\subsection{Future-Aligned Proposal Prediction}
\label{sec:formulation}

We formulate research proposal generation as a time-sliced learning problem over scientific papers. Let $\mathcal{C}$ denote a time-stamped corpus, where each paper $p$ has publication time $t_p$. For a target paper $Y$ published after cutoff time $t_C$, papers with $t_p \le t_C$ are treated as observable context, while papers with $t_p > t_C$ form a held-out future corpus.

For each target paper $Y$, we construct an input instance $(q,S)$, where $q$ is a research question derived from $Y$ and $S$ is a set of inspiring papers selected from its references. Given $(q,S)$, the model generates a structured proposal $\hat{P}$:
$(q,S) \rightarrow \hat{P}$.

Rather than reconstructing $Y$, the goal is to generate a proposal that anticipates research directions later realized in the future literature.

We represent $\hat{P}$ using a structured schema consisting of \textit{Research Question}, \textit{Hypothesis}, \textit{Proposed Method}, \textit{Novelty Claims}, and \textit{Experimental Details}, enabling both complete proposal generation and fine-grained evaluation.

\subsection{Verifiable Learning Objective}

Directly optimizing subjective notions such as novelty or creativity is difficult to scale, so we use \emph{future alignment} as a verifiable surrogate objective. Let
\[
\mathcal{C}_{\text{future}} = \{p \in \mathcal{C} \mid t_p > t_C\}
\]
denote the future corpus. Given a generated proposal $\hat{P}$, we first retrieve a candidate set of future papers using embedding similarity. Let $e(\cdot)$ denote the embedding model, and let $\mathcal{R}_k(\hat{P})$ denote the top-$k$ retrieved future papers for $\hat{P}$:
\[
\mathcal{R}_k(\hat{P})
=
\operatorname{TopK}_{p \in \mathcal{C}_{\text{future}}}
\cos\!\bigl(e(\hat{P}), e(p)\bigr).
\]
We then score the semantic alignment between $\hat{P}$ and each retrieved paper $p$ using an LLM judge, denoted by $s_{\text{llm}}(\hat{P}, p)$. The \textbf{Future Alignment Score (FAS)} is defined as
\[
\mathrm{FAS}(\hat{P})
=
\max_{p \in \mathcal{R}_k(\hat{P})}
s_{\text{llm}}(\hat{P}, p).
\]

FAS measures whether a proposal aligns strongly with at least one plausible future research direction. We frame FAS as a \emph{verifiable surrogate} in the narrow sense: it provides a scalable, automatic signal grounded in publication outcomes, not a complete measure of proposal quality (see Limitations and Appendix~\ref{sec:fas_discussion} for extended discussion). The $\max$ aggregation is motivated in Appendix~\ref{sec:fas_discussion}. Since proposals are structured, we also compute component-level FAS for fine-grained analysis of hypotheses, methods, novelty claims, and experimental plans. Because FAS is not directly differentiable, in Section~\ref{sec:supervision} we construct time-consistent supervision from historical papers as a tractable proxy for this objective.

\paragraph{Component-level Future Alignment}
Since proposals are structured, we also compute alignment separately for each component, which enables fine-grained analysis of whether a model better anticipates future hypotheses, methods, novelty claims, or experimental plans.

\vspace{-1mm}
\subsection{Time-Consistent Supervision from Historical Papers}
\label{sec:supervision}
To optimize for the FAS objective, we construct supervision from historical papers by converting published papers together with their cited previous papers into time-consistent training data. Given a target paper $Y$ and cutoff time $t_C$, we construct
\[
q = g_q(Y), \quad S = g_S(Y, t_C), \quad \tilde{P} = g_P(Y, S),
\]
where $q$ is a leakage-controlled research question, $S$ is a set of historically available inspiring papers, and $\tilde{P}$ is a structured proposal target consistent with that context.

\paragraph{Research Question Extraction.}
We derive $q$ from $Y$ using a prompt that suppresses leakage of key ideas and details, so that $q$ resembles a plausible pre-publication problem statement.

\paragraph{Inspiring Paper Construction}
Let $\mathrm{Refs}(Y)$ denote the references of $Y$. We construct $S$ with a coarse-to-fine procedure: a lightweight heuristic first selects a shortlist $S_{\mathrm{top}}(Y)$, and an LLM-based selector then chooses the final inspiring papers,
\[
S = g_{\mathrm{LLM}}(Y, S_{\mathrm{top}}(Y)),
\]
favoring citations likely to have served as genuine inspirations rather than peripheral references.

\paragraph{Proposal Target Synthesis}
Given $(q,S)$, we synthesize a forward-looking structured proposal $\tilde{P}$ containing a hypothesis, method, novelty claims, and experimental plan, yielding supervised pairs $(q,S)\rightarrow\tilde{P}$ that provide scalable supervision.

\vspace{-1mm}
\subsection{Citation-Grounded Reasoning Traces}

Reasoning is a core part of scientific ideation: human researchers form proposals by analyzing gaps in prior work, borrowing and recombining ideas from existing papers, and refining these insights into methods and experiments. Supervising only the final proposal therefore misses an important part of the learning problem. To better capture this process, we synthesize \emph{citation-grounded reasoning traces} that model how inspiring papers $S$ are transformed into a new proposal through gap identification and inspiration borrowing.

We further introduce \textbf{Stepwise CoT SFT}, a more fine-grained form of reasoning supervision. Instead of placing reasoning in a single monolithic block, Stepwise CoT interleaves reasoning with proposal construction in three stages: (1) gap analysis and inspiration borrowing for \emph{research question and hypothesis} generation, (2) \emph{method reasoning} for the \emph{proposed method}, and (3) \emph{experimental planning} for \emph{experimental details}. This formulation teaches proposal generation as staged scientific planning rather than one-shot text continuation, and forms the main reasoning component of our framework.

For comparison, we also consider two simpler supervision formats. In \textbf{CoT SFT}, the entire reasoning trace is placed before the proposal as a single block: $
(q,S) \mapsto (R, \tilde{P})$,
where $R$ contains gap analysis and inspiration borrowing. In \textbf{Direct SFT}, the model is trained to generate structured proposals without intermediate reasoning traces.

All reasoning traces are synthesized using a strong reasoning language model, and formatting details and prompt templates are provided in Appendix~\ref{sec:reasoning_details}.

\begin{table*}[t]
\centering
\caption{
Main results (FAS; higher is better) for future-aligned proposal prediction.
We report component-wise FAS along with the overall FAS. Future-aligned SFT improves the FAS substantially over the unaligned baselines, while synthetic structured reasoning trace supervision provides additional gains, improving the overall score by 10.6\%.
}
\vspace{-2mm}
\label{tab:main_results}
\resizebox{0.94\textwidth}{!}{%
\begin{tabular}{lccccccc}
\toprule[1pt]
\textbf{Method} & \textbf{Hypothesis} & \textbf{Proposed Method} & \textbf{Novelty Claims} & \textbf{Exp. Details}  & \textbf{Overall} \\
\midrule[0.5pt]
\rowcolor{lightgray}
\multicolumn{6}{c}{\textbf{Llama-3.1-8B-Instruct}} \\ 
\midrule[0.5pt]
RQ only & 63.0 & 52.8 & 51.1 & 52.4 & 60.0 \\
Paper only & 55.2 & 49.4 & 46.9 & 48.0  & 52.4 \\
Prompting & 64.5 & 56.8 & 54.4 & 55.0 & 62.1 \\
AI-Researcher & 57.9 & 46.0 & 45.1 & 44.4 & 53.7  \\
Chain-of-Ideas & 63.2 & 54.1 & 51.6 & 45.9 & 59.3  \\
\textbf{Ours (Future-aligned SFT + Stepwise CoT)} & \textbf{68.1} & \textbf{61.0} & \textbf{58.7} & 51.9 & \textbf{65.3} \\

\quad {w/o reasoning traces} & 66.7 & 58.3 & 56.9 & 54.6 & 64.1 \\ 

\quad {w/o stepwise} & 67.2 & 59.2 & 57.6 & \textbf{55.1} & 65.0 \\

\midrule[0.5pt]
\rowcolor{lightgray}
\multicolumn{6}{c}{\textbf{Qwen2.5-7B-Instruct}} \\ 
\midrule[0.5pt]
RQ only & 66.7 & 56.6 & 55.2 & 52.0 & 63.3 \\
Paper only & 56.4 & 50.5 & 46.6 & 47.2 & 52.9 \\
Prompting & 66.3 & 57.8 & 55.1 & 54.8 & 63.6 \\
AI-Researcher & 58.0 & 47.5 & 45.6 & 44.8 & 54.1   \\
Chain-of-Ideas & 62.0 & 53.2 & 50.3 & 47.1 & 59.3 \\
\addlinespace[2pt]
\textbf{Ours (Future-aligned SFT + Stepwise CoT)} 
& \textbf{68.7} & \textbf{60.5} & 59.1 & \textbf{54.9} & \textbf{66.5} \\

\quad w/o reasoning traces
& 67.9 & 60.0 & 58.7 & 54.9 & 65.3 \\
\quad w/o stepwise
& 68.3 & 60.2 & \textbf{59.2} & 54.8 & 65.9 \\

\midrule[0.5pt]
\rowcolor{lightgray}
\multicolumn{6}{c}{\textbf{Qwen2.5-14B-Instruct}} \\ 
\midrule[0.5pt]
RQ only & 65.8 & 56.4 & 54.4 & 53.6 & 62.6\\
Paper only & 54.3 & 49.3 & 45.6 & 46.6 & 51.3 \\
Prompting & 65.8 & 57.1 & 55.1 & 55.6 & 63.0  \\
AI-Researcher & 58.5 & 48.1 & 46.0 & 45.0 & 55.3  \\
Chain-of-Ideas & 63.8 & 54.7 & 52.1 & 46.8 & 60.8 \\
\addlinespace[2pt]
\textbf{Ours (Future-aligned SFT + Stepwise CoT)} 
& \textbf{71.4} & \textbf{63.5} & \textbf{61.8} & 56.7 & \textbf{69.7} \\

\quad w/o reasoning traces
& 68.0 & 60.0 & 58.9 & 55.1 & 65.1 \\
\quad w/o stepwise
& 68.7 & 60.0 & 58.2 & \textbf{56.8} & 66.1 \\
\bottomrule[1pt]
\end{tabular}
}
\vspace{-3mm}
\end{table*}

\vspace{-2mm}
\section{Experiments}
\vspace{-1mm}
\subsection{Experimental Setup}

\paragraph{Corpus and Temporal Split}
Our corpus consists of papers from major machine learning venues (NeurIPS, ICML, and ICLR). Papers from 2024 are used to construct training supervision, while papers from 2025 serve as future evaluation targets. We randomly sample 2,823 training instances and 819 evaluation instances. Each instance pairs one target paper with up to five inspiring papers from its pre-cutoff citations, yielding 21{,}835 paper occurrences.

\vspace{-1mm}
\paragraph{Evaluation}
We encode generated proposals and future papers using text-embedding-3-large~\citep{openai_text_embedding_3_large_2024}, retrieve the top-$k$ future candidates ($k=10$), and use GPT-4.1-mini~\citep{openai_gpt41_2025} as the semantic judge for Future Alignment Score (1-10 scale with detailed rubrics). We show that the evaluation is robust to retrieval depth $k$, embedding models, and LLM judge selection (inter-judge agreement statistics across two additional judges GPT-4o-mini and GPT-5-mini) in Appendix~\ref{sec:fas_robustness}.
\vspace{-0.5mm}
\paragraph{Models and Training Setup}
We evaluate Qwen2.5-7B-Instruct, Qwen2.5-14B-Instruct~\citep{qwen2.5}, and Llama-3.1-8B-Instruct~\citep{grattafiori2024llama} under both prompting and supervised fine-tuning regimes. For supervised training, we fine-tune each model on the synthesized proposal supervision using LoRA adapters~\citep{hu2022lora}. We use a generation temperature of 0.7 uniformly for all systems at inference time.

\vspace{-2mm}
\subsection{Baselines}
\vspace{-1mm}
\paragraph{Prompting Baselines}
We consider three input configurations: 
(1) Research Question Only: the model receives only $q$.  
(2) Papers Only: the model receives only the inspiring papers $S$.  
(3) Research Question + Papers: the full input $(q, S)$.

\vspace{-1mm}
\paragraph{Baselines from Prior Paradigms}
Since future-aligned proposal prediction is a new task formulation, there is no existing method designed for the same input/output setting. We therefore include adapted baselines that represent the closest prior paradigms:

(1) \textbf{AI-Researcher}~\citep{si2025can}: generates research proposals by first producing several seed research ideas grounded in retrieved literature, expanding them into full proposals, then using an LLM ranker to rank the proposals.

(2) \textbf{Chain-of-Ideas (CoI)}~\citep{li-etal-2025-chain-ideas}: CoI models the evolution of research ideas through chains of related papers and predicts future research directions before generating a candidate idea and experimental design. 


\vspace{-1mm}
\paragraph{Supervised Training Regimes}
Our main supervised training method is \textbf{Future-aligned SFT + Stepwise CoT}, which fine-tunes the model on synthesized proposals augmented with interleaved reasoning traces aligned to proposal components. To isolate the contribution of reasoning supervision, we also consider two ablated variants: \textbf{w/o reasoning traces}, which removes intermediate reasoning and trains only on structured proposal targets (Direct SFT), and \textbf{w/o stepwise}, which retains reasoning supervision but places it in a single block before the proposal rather than distributing it across the generation process (CoT SFT).

\vspace{-2mm}
\subsection{Main Results}
\vspace{-1mm}
\label{sec:results}
As shown in Table~\ref{tab:main_results}, our full method achieves the best overall FAS across all three backbones. Relative to standard prompting with the same base model, the gains are consistent on Llama-3.1-8B (+5.2\% overall FAS), Qwen2.5-7B (+4.6\%), and especially Qwen2.5-14B (+10.6\%), indicating that future-aligned supervision is effective across scales and that larger models benefit most from the additional training signal. These results indicate that future-aligned supervision substantially improves proposal generation relative to inference-time prompting alone. Statistical significance tests for these FAS comparisons are reported in Appendix~\ref{sec:fas_significance}. Appendix~\ref{sec:qualitative} presents qualitative case studies that illustrate the proposal differences underlying these FAS improvements.

The ablations further clarify where these gains come from. Removing reasoning traces (\textit{w/o reasoning traces}) already reduces performance on all three backbones, and replacing stepwise reasoning with a single monolithic reasoning block (\textit{w/o stepwise}) leads to a further drop in FAS. This pattern is strongest on Qwen2.5-14B, where the full method reaches 69.7 overall FAS, compared with 66.1 for \textit{w/o stepwise} and 65.1 for \textit{w/o reasoning traces}. The component-level results show that the gains are concentrated on \emph{Hypothesis} and \emph{Proposed Method}, while \emph{Experimental Details} improves less consistently. For example, on Qwen2.5-14B, the full method improves Hypothesis from 68.0 to 71.4 and Proposed Method from 60.0 to 63.5 relative to \textit{w/o reasoning traces}, whereas Experimental Details changes more modestly (55.1 to 56.7). This suggests that citation-grounded stepwise reasoning mainly helps the reasoning-intensive stages of proposal generation, especially problem formulation and method design. While improvements are consistent, gains on experimental design are more modest, suggesting that planning realistic experiments remains challenging.

\vspace{-0.5mm}
We also compare against adapted baselines from prior ideation paradigms, including AI-Researcher and Chain-of-Ideas. These methods consistently underperform direct prompting under our future-alignment metric. On Qwen2.5-14B, for instance, AI-Researcher reaches 55.3 overall FAS, and Chain-of-Ideas reaches 60.8, both below standard prompting (63.0) and well below our full method (69.7). We stress, however, that these systems are designed for broader open-ended ideation rather than the forecasting-style objective studied here. Their lower FAS should therefore be interpreted specifically with respect to future alignment, not as a universal judgment of proposal quality. Nevertheless, the comparison shows that performance gains in our setting come from future-aligned supervision rather than from more elaborate prompting workflows alone.

Finally, the input ablations highlight the importance of problem specification. Using only the research question remains reasonably competitive, while using only inspiring papers is consistently much weaker across all models. On Qwen2.5-14B, for example, RQ-only achieves 62.6 overall FAS, close to full prompting at 63.0, whereas Paper-only drops to 51.3. This suggests that the research question provides the main high-level constraint for proposal generation, while inspiring papers contribute complementary information that becomes most useful when the model is trained to exploit them through future-aligned supervision and citation-grounded reasoning.

\begin{figure}[t]
    \centering
    \includegraphics[width=0.98\linewidth]{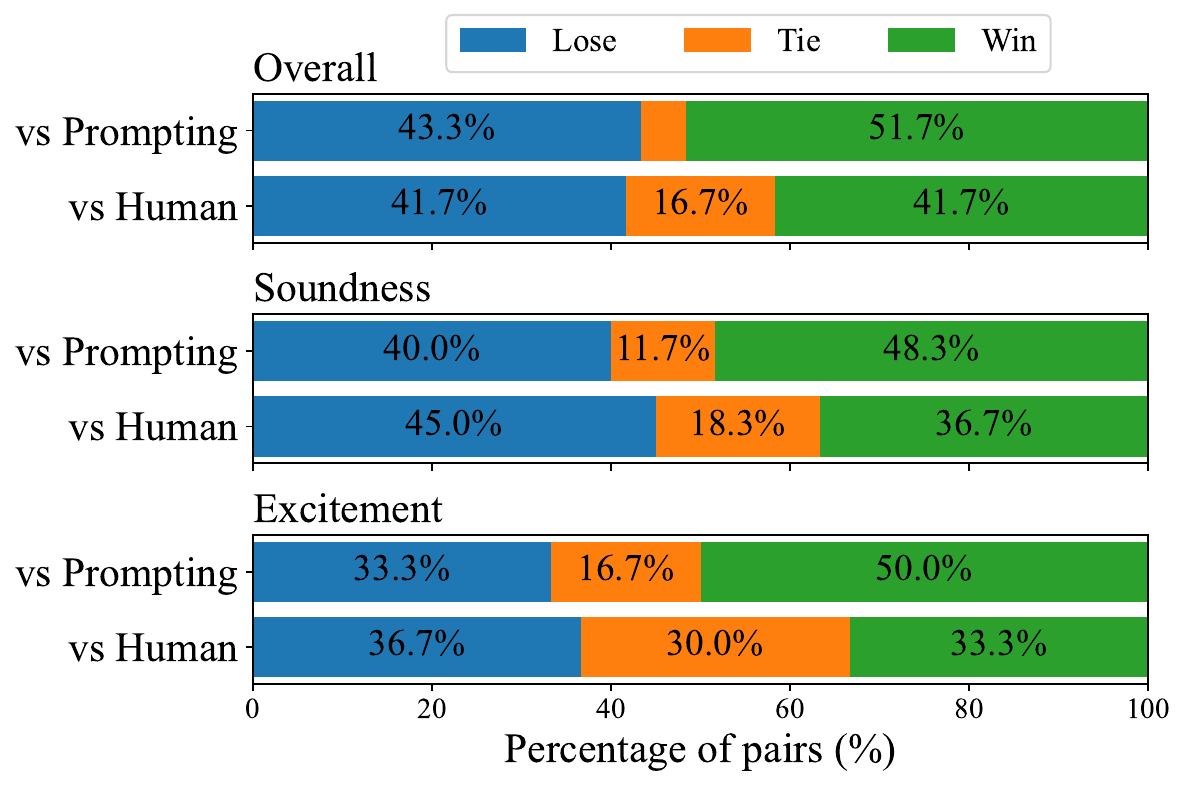}
    \vspace{-2mm}
    \caption{Pairwise human evaluation results (win/tie/lose). Each stacked bar shows the fraction of instances where Stepwise CoT is preferred (win), the two proposals are judged equivalent (tie), or Stepwise CoT is not preferred (lose), aggregated by majority vote across three annotators.
    }
    \label{fig:human_eval}
    \vspace{-4mm}
\end{figure}

\begin{figure*}[t]
    \centering
    \includegraphics[width=0.98\linewidth]{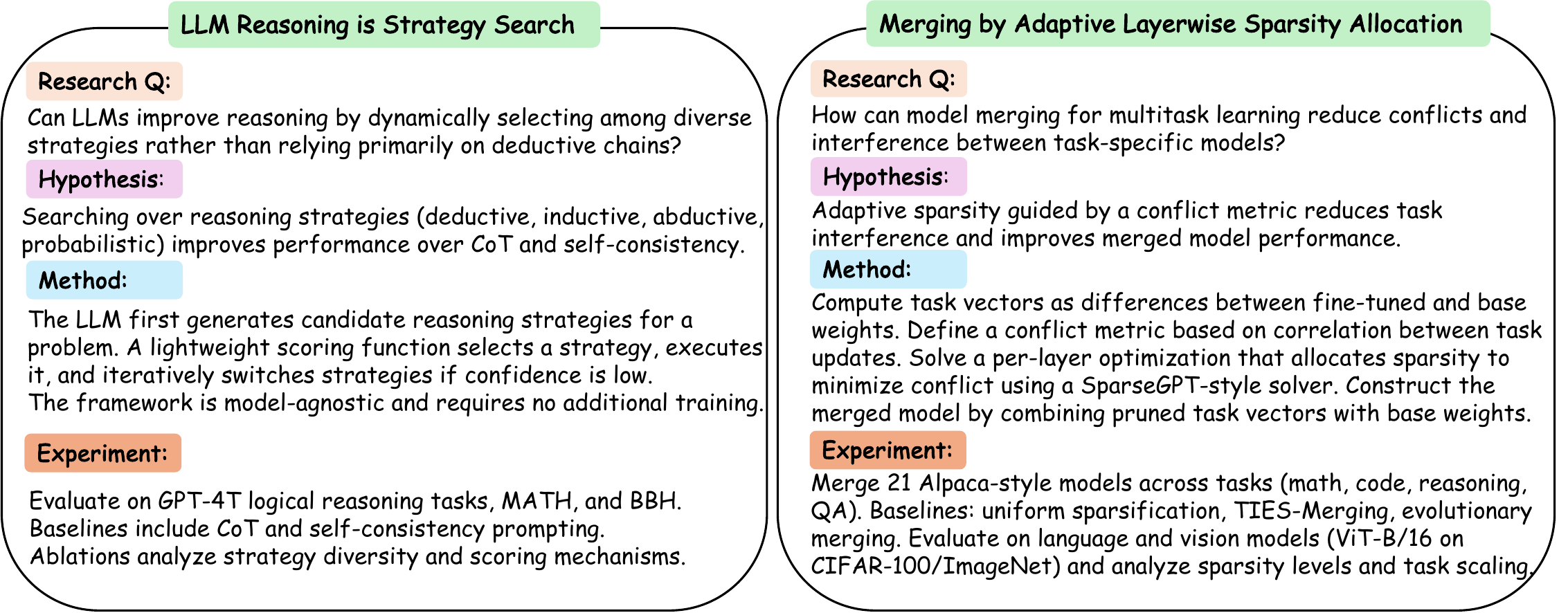}
    \vspace{-2mm}
    \caption{Two proposals generated by Qwen2.5-14B- Instruct (stepwise CoT tuned). The content is summarized for readability. The proposals are textually sound and are turned into reasonable experimental results and findings with the implementation and execution of code agents.}
    \label{fig:proposal}
    \vspace{-3mm}
\end{figure*}

\vspace{-2mm}
\subsection{Human Evaluation}
\vspace{-1mm}
\label{sec:human_eval}
To examine whether improvements in Future Alignment Score (FAS) correspond to stronger proposals under expert judgment, we conduct a pairwise human study comparing proposals generated by our Qwen2.5-14B Stepwise CoT model against (i) human-derived proposals from published papers and (ii) prompting-only proposals. Each pair is written for the same research question, and all proposals are in the format defined in Section~\ref{sec:formulation}. Annotators evaluate each pair along three dimensions: soundness, excitement, and overall assessment. We annotate 120 pairs in total (60 per comparison), each judged by three domain-expert graduate students with prior conference reviewing experience; judgments are aggregated by majority vote.

As shown in Figure~\ref{fig:human_eval}, Stepwise CoT is competitive with human-derived proposals overall (25 wins / 10 ties / 25 losses). This suggests that proposals with high future alignment can also be judged favorably by human experts, supporting FAS as a meaningful---though not exhaustive---surrogate for proposal quality. This comparison also exhibits a relatively high tie rate, especially for excitement (18 ties out of 60), and low unanimity (5--12\%), indicating that distinctions between strong model-generated and human-derived proposals can be subtle even for expert annotators.

Compared to prompting-only proposals, Stepwise CoT is preferred more often across all dimensions. This mirrors the gains observed under FAS and provides complementary evidence that higher future alignment is associated with improvements in expert-perceived proposal quality, rather than merely better matching to the future corpus. This comparison also shows substantially higher unanimity (26.7--31.7\%), indicating more consistent annotator preferences when distinguishing Stepwise CoT from prompting baselines.

\vspace{-1mm}
\subsection{Executable Proposal Case Studies}

To assess whether generated proposals correspond to executable research ideas, we implement two high-FAS proposals produced by our best model. We provide the machine-generated proposals directly to a code agent, which follows the proposed methodology with minor human assistance. Full implementation details and results are provided in Appendix~\ref{sec:implementation}.

\vspace{-2mm}

\paragraph{Prompting Strategy Proposal} introduces a prompting technique\footnote{Because the model only has access to papers up to 2024, novelty here is defined relative to research appearing in 2025.} based on a meta strategy selector over deductive, inductive, abductive, and enumerative reasoning. Rather than relying on a single reasoning chain, it explores multiple strategies and selects an answer by agreement-based scoring. This combination goes beyond the prompting methods available in the input context, and improves MATH accuracy by 4.17\% over baselines.

\vspace{-2mm}
\paragraph{Model Merging Proposal} introduces MALS (Merging by Adaptive Layerwise Sparsity Allocation), a model merging method that addresses conflicts in task-vector sparsification. Instead of using uniform sparsification, MALS estimates layerwise conflict from cross-task correlation and sign disagreement, then allocates sparsity adaptively across layers. This conflict-aware design goes beyond existing heuristic sparsification strategies and yields strong reasoning performance on Mistral-7B.

These case studies provide qualitative evidence that optimizing for future alignment can produce proposals that are practically implementable. More importantly, it indicates the potential that our model can be incorporated into a fully automated end-to-end research assistant system.

\vspace{-2mm}
\section{Analysis}
\vspace{-1mm}

\begin{table}[t]
\centering
\caption{Citation-type ablation: $\Delta$ FAS (Ablation $-$ Baseline) by proposal component.}
\vspace{-2mm}
\label{tab:citation_ablation}
\resizebox{0.48\textwidth}{!}{%
\begin{tabular}{l c c c c}
\toprule[1pt]
\textbf{Ablation Type} & \textbf{Hyp.} & \textbf{Method} & \textbf{Novelty} & \textbf{Overall} \\
\midrule[0.5pt]
background ($n$=775) & $-$6.75 & $-$7.21 & $-$8.31 & $-$6.59 \\
method ($n$=765)   & $-$\textbf{6.93} & $-$\textbf{7.37} & $-$\textbf{8.35}  & $-$\textbf{6.63} \\
benchmark ($n$=248) & $-$6.41 & $-$6.85 & $-$7.30  & $-$6.13 \\
\bottomrule[1pt]
\end{tabular}
}
\vspace{-4mm}

\end{table}

\subsection{Citation Sensitivity by Citation Type}
\label{subsec:citation_ablation}

To understand whether the model genuinely leverages inspiring papers rather than generating generic proposals, we perform a citation sensitivity analysis using type-aware ablations. We categorize each inspiring citation in $S$ into three coarse types based on its role in the original paper: \emph{Foundational/Background} (theoretical context and problem framing), \emph{Method/Technical} (inspiring techniques or components), and \emph{Benchmark/Experimental} (datasets, evaluation metrics, or experimental protocols).

We label inspiring papers for all 819 test instances (4{,}084 papers total) using GPT-4.1-mini. Papers may receive multiple labels, and 43.5\% fall into more than one category. Background and method citations are common (72.1\% and 62.8\% of inspiring papers, respectively), while benchmark citations are relatively rare (9.2\%).

For each instance, we generate a proposal $\hat{P}$ using our best-performing model with the full inspiring set $S$, and then remove all citations of a given type $t$ before regenerating:
\[
\hat{P}^{(-t)} = f\big(q,\, S \setminus S^{(t)}\big),
\]

where $S^{(t)} \subseteq S$ denotes the subset of inspiring papers labeled with type $t$. Sensitivity is measured by the drop in Future Alignment Score (FAS). We report results only on instances where the removed citation type is present.

As shown in Table~\ref{tab:citation_ablation}, removing either background or method citations leads to a similar degradation in overall FAS (both around a $9.6\%$ drop), suggesting that contextual framing and technical inspiration contribute comparably to forecasting future research directions. At the component level, \emph{Novelty Claims} are the most sensitive to citation removal across all citation types, indicating that inspiring papers help the model articulate distinctive contributions relative to prior work. Removing benchmark citations has the smallest effect on overall alignment (an $8.9\%$ drop on instances containing such citations), suggesting that conceptual framing and methodological inspiration play a larger role than explicit benchmark references in anticipating future research trajectories.

\begin{table}[t]
\centering
\caption{Multi-dimensional LLM judge evaluation (1--5 scale). Future-aligned tuned models consistently outperform the baselines, while the Stepwise-CoT model achieves the highest scores on all three dimensions.}
\vspace{-2mm}

\label{tab:llm_judge}
\resizebox{0.48\textwidth}{!}{%
\begin{tabular}{lcccc}
\toprule[1pt]
\textbf{Model} & \textbf{Resource} & \textbf{Task--Method} & \textbf{Task--Exp.} & \textbf{Avg.} \\
\midrule[0.5pt]
Prompting & 3.26 & 3.20 & 2.99 & 3.15 \\
CoI & 3.52 & 3.20 & 3.12 & 3.28 \\
AI-Researcher & 3.30 & 3.13 & 3.03 & 3.16 \\
\addlinespace[2pt]
{\textbf{Future-aligned SFT}} & \textbf{3.63} & \textbf{3.42} & \textbf{3.21} & \textbf{3.42} \\
\quad {\small w/o reasoning traces} & 3.54 & 3.41 & 3.16 & 3.37 \\
\quad {\small w/o stepwise} & 3.39 & 3.30 & 3.07 & 3.25 \\

\bottomrule[1pt]
\end{tabular}
}
\vspace{-4mm}
\end{table}

\subsection{Multi-Dimensional LLM-Based Judging}
\label{subsec:llm_judge}

While FAS measures predictive alignment with future research, it does not directly assess the structural quality and feasibility of generated proposals.
We therefore conduct an additional evaluation using an LLM-based judge across three dimensions: \emph{Resource Validity}, \emph{Task--Method Consistency}, and \emph{Task--Experiment Consistency}.
Each dimension is scored on a 1--5 scale (higher is better). The full LLM judge prompt is provided in Appendix~\ref{sec:ana_llm_judge}.

As illustrated in Table~\ref{tab:llm_judge}, the Stepwise-CoT model achieves the highest scores on all three dimensions, indicating that structured reasoning supervision improves not only the coherence of generated proposals but also the validity of required resources and the consistency between proposed tasks, methods, and evaluation protocols.

Across models, \emph{Task--Experiment Consistency} receives the lowest scores (approximately $3.0$--$3.2$), suggesting that designing experiments that properly evaluate the proposed task remains challenging for current language models.
Nevertheless, all future-aligned fine-tuned models outperform baseline approaches such as Chain-of-Ideas and AI-Researcher, demonstrating that future-aligned supervision improves both proposal feasibility and methodological consistency.

\vspace{-2mm}
\section{Related Work}
\vspace{-1mm}
\paragraph{LLM for Research Ideation and Scientific Assistance}
Recent work has explored the use of large language models as research assistants for accelerating scientific discovery in various domains. Many works focus on partial processes in scientific discovery, such as idea generation~\citep{wang-etal-2024-scimon,chen2025beyond,si2025can,10.1145/3706598.3714057,xu2026idea2story,baek-etal-2025-researchagent,10.1145/3711896.3737419,kumar-etal-2025-large}, literature synthesis~\citep{wang-etal-2019-paperrobot,wang2024autosurvey}, idea implementation~\citep{jansen-etal-2025-codescientist,si2026the,si2026towards}, and paper reviewing~\citep{d2024marg,liang2024can,zhu-etal-2025-deepreview}. Another line of work focuses on end-to-end scientific workflow, such as building multi-agent systems to simulate scientific workflow~\citep{yu2025researchtown} or automate the research pipeline to produce a full research paper~\citep{lu2024ai,schmidgall2025agentlaboratoryusingllm,schmidgall2025agentrxiv,gottweis2025towards,yamada2025ai,miyai2026jr}. Benchmarks have also been created to evaluate research agents on research tasks (e.g. research coding)~\citep{tian2024scicode,huang2024mlagentbench,nathani2025mlgym,chan2025mlebench,weng2025opendiscovery}.
Our work instead focuses on scalable training and evaluation for proposal generation.

\vspace{-1mm}

\paragraph{Scientific Forecasting and Predictive Evaluation}
A growing body of work studies whether language models can predict future scientific outcomes, trends, or empirical results. For example, \citet{luo2025large} use LLMs to predict neuroscience results. \citet{wen2025predicting} investigate forecasting empirical AI research outcomes with LLMs, while other datasets analyze innovation patterns and scientific reasoning from historical research trajectories \citep{liu2026sci}. Other works also try to predict trending research topics~\citep{gu2024impactcast,liu2026sci} or generate the future works section of a paper~\citep{al2025futuregen}. Recently, \citet{ajith2026prescience} create a benchmark to evaluate scientific forecasting across the entire research workflow. Our work is inspired by this line of research and introduces the future alignment objective for proposal generation.

\vspace{-2mm}
\section{Conclusion}
\vspace{-1mm}

We introduce future-aligned research proposal prediction, a formulation that evaluates generated proposals by whether they anticipate research directions appearing in future publications. We construct a time-consistent dataset derived from published papers and their citations, and propose the Future Alignment Score (FAS) to measure semantic alignment between generated proposals and a held-out future corpus. Experiments show that future-aligned supervised fine-tuning substantially improves predictive alignment over baselines, with structured reasoning supervision providing additional gains. Complementary evaluations with LLM-based judging, human comparisons, and executable case studies suggest that the resulting proposals are coherent, feasible, and practically implementable. These results highlight the potential of time-grounded evaluation signals for training language models to better assist scientific exploration.

\section*{Limitations}
\paragraph{Limited Evaluation Domain and Horizon} The dataset and evaluation are limited to machine learning papers from NeurIPS, ICML, and ICLR, and use a one-year temporal split (2024$\to$2025) constrained by the knowledge cutoffs of our backbone models. It is unclear how the framework generalizes to domains with different citation cultures or slower publication cycles, or to longer-term and more disruptive research shifts. We note, however, that the framework itself is not ML-specific—it requires only a time-stamped corpus and citation graph, both available in many other fields (e.g., bioRxiv, ACL Anthology, arXiv categories)—and extending to these settings is a natural direction for future work.

\paragraph{Biased Objective} FAS measures similarity to future published directions and, by construction, rewards mainstream or incremental research directions that dominate top-venue proceedings. It does not directly capture novelty or scientific correctness, and a strong proposal that is original but not realized in the future literature can be unfairly scored poorly. Publication bias is not unique to FAS: any retrospective signal in science inherits similar confounds. We mitigate this by complementing FAS with three independent signals not driven by publication popularity—expert human evaluation (Section~\ref{sec:human_eval}), multi-dimensional LLM judging (Section~\ref{subsec:llm_judge}), and executable case studies (Section~\ref{sec:implementation})—and we frame FAS as a verifiable surrogate rather than a complete measure of research quality.

\paragraph{Data Synthesis and Supervision} Many steps in the data synthesis process involve LLM usage, such as inspiring paper selection and reasoning trace generation, which could introduce model bias and errors. Appendix~\ref{sec:extended_discussion} provides extended discussion of these and other design choices.

\section*{Ethical Statement}
A potential risk of proposal-generation systems is dual use: models that can draft plausible research plans could be misused to accelerate harmful or unethical research. Our experiments focus on mainstream machine learning topics and do not target domains that are inherently hazardous. We also emphasize that future alignment is not equivalent to novelty, correctness, or societal benefit; proposals that align with future publications may still be unhelpful or inappropriate, and proposals that do not align may still be valuable. Any deployment of such systems should include human oversight, domain-specific safety checks, and additional review for harmful content or unsafe experimental recommendations.

Moreover, our evaluation relies in part on LLM-based judges and tool-assisted verification, which may reflect model biases or errors. We mitigate this by using fixed prompts and deterministic settings where possible, reporting uncertainty through confidence intervals in human studies, and treating the tool-assisted checks as analysis rather than definitive ground truth.

\section*{Acknowledgements}
The authors would like to thank Hui Ren, Xuejun Zhang, Jiateng Liu, Jeonghwan Kim, Yanjun Zhao, Bingxuan Li, Xueqiang Xu, Haojin Wang, and Chenyu Li for helpful discussions and data annotation.
This research is based upon work supported by NSF Molecule Maker Lab Institute, an AI Institute for Molecular Discovery, Synthesis Strategy, and Manufacturing funded by the U.S. National Science, Foundation under Awards No. 2019897 and 2505932, NSF NAIRR award, DOE Center for Advanced Bioenergy and Bioproducts Innovation U.S. Department of Energy, Office of Science, Office of Biological and Environmental Research under Award Number DESC0018420. The views and conclusions contained herein are those of the authors and should not be interpreted as necessarily representing the official policies, either expressed or implied, of NSF, DoE, or the U.S. Government. The U.S. Government is authorized to reproduce and distribute reprints for governmental purposes notwithstanding any copyright annotation therein. This research used the Delta and DeltaAI advanced computing and data resources, which are supported by the National Science Foundation (award OAC 2320345 and award OAC 2005572) and the State of Illinois. Delta and DeltaAI are joint efforts of the University of Illinois Urbana-Champaign and its National Center for Supercomputing Applications.

\bibliography{custom}

\clearpage
\appendix

\section{Extended Discussion}
\label{sec:extended_discussion}

\subsection{Scope and Methodological Considerations}
\label{sec:scope_discussion}

\paragraph{Scope of the Time-Sliced Forecasting Framing.}
Our framing fits the dominant mode of scientific progress—incremental and recombinative work built on recent literature—but is fundamentally less suited to \emph{transformative} proposals whose value lies in unpredictability. Ideas that anticipate paradigm shifts or draw on unusual cross-disciplinary connections cannot be captured by alignment with imminent future publications, since their merit is invisible until the field reorganizes around them. Generating such proposals likely requires complementary approaches (e.g., counterfactual or open-ended exploration) beyond the scope of this work.

\paragraph{Domain-Specific Time Horizons.}
We evaluate on AI/CS, where a one-year horizon ($2024{\to}2025$) captures substantial follow-up activity. In slower-moving fields such as biology, mathematics, or theoretical physics, where landmark ideas can take five or more years to propagate, this horizon would be too short and FAS would systematically underestimate proposal quality. Applying the framework to such domains requires recalibrating the time slide to the natural pace of the field, together with corresponding adjustments to backbone knowledge cutoffs and supervision construction.

\paragraph{Does the Framework Incentivize Incremental Research?}
A natural concern is that optimizing for future-publication alignment implicitly rewards generic, hot-topic proposals. The $\max$ aggregation partially mitigates this by rewarding match to a single specific future paper rather than the centroid of a topical neighborhood, and our human and multi-dimensional LLM evaluations assess novelty directly as a counterweight. Nonetheless, models optimized purely for FAS will in equilibrium be drawn toward mainstream directions. We see FAS as appropriate for assistants exploring the modal kind of research—incremental, community-validated, recombinative—and as inappropriate as a sole objective for systems targeting contrarian or transformative ideas. The two objectives are complementary; we recommend pairing future-alignment training with novelty- or diversity-oriented objectives in deployment.

\paragraph{Dimension-Specific Predictability.}
Different proposal aspects are likely predictable to different degrees. Methods and experimental setups, which inherit heavily from recent literature, may forecast relatively well; hypotheses and high-level research questions depend more on emerging empirical findings and are likely harder; novelty claims sit at the extreme end, since the most novel proposals are by definition those least anticipated. Our component-level FAS analysis (Section~\ref{sec:results}) begins to disentangle these dimensions, and a more systematic study—varying horizon, component, and domain—would shed light on which aspects of scientific creativity are learnable from history. We leave this to future work.

\subsection{Design Choices in FAS}
\label{sec:fas_discussion}

\paragraph{Why Max Aggregation.}
Research proposal generation is a one-of-many prediction problem: a strong proposal needs to anticipate at least one plausible future direction, not the average of all directions in a topical neighborhood. A proposal that closely matches a single future paper represents a successful forecast even when other top-$k$ retrieved papers are loosely related. Mean and top-$k$-mean aggregations would instead reward proposals that sit close to the centroid of the retrieved neighborhood, biasing the metric toward generic, broadly-topical proposals.

\subsection{Data Construction Rationale}
\label{sec:data_discussion}

\paragraph{Inspiring Papers at Training vs. Inference Time.}
The citation-based selection of $S$ is a training-data construction step that approximates the curated inputs a researcher would naturally provide at inference; it is not a claim of foresight. At test time the model receives whatever inspiring papers the user provides, with no access to future information. When pre-curation is infeasible, the framework can be paired with a retrieval module over the pre-cutoff corpus (as in AI-Researcher and Chain-of-Ideas).

\paragraph{Choice of $|S|$.}
We fix $|S|{=}5$ to balance contextual grounding against prompt length—the mean training prompt already reaches 1{,}755 words (Section~\ref{sec:proposal_statistics}). A typical ML paper draws its core methodological inspiration from a handful of references, which the two-stage selection (Appendix~\ref{sec:data_details}) surfaces. Studying sensitivity to $|S|$ is left to future work.

\section{Details of Implemented Proposals}
\label{sec:implementation}

We present the detailed implementation and experiment results of the two generated proposals. We use Cursor with Claude-4.5-Opus for the implementation. We also include the LLM-generated analysis over the results to demonstrate the potential to build an end-to-end framework that generates the full paper.

\subsection{Strategy Search}
\paragraph{Implementation} We implement Strategy Search, a multi-strategy reasoning framework where the LLM explicitly proposes, executes, and scores multiple reasoning strategies before selecting the best answer. Unlike standard Chain-of-Thought (CoT), which commits to a single deductive reasoning chain, Strategy Search explores four distinct reasoning modes:

Deductive: Apply logical rules step-by-step from premises to conclusion (best for formal proofs, syllogisms)
Inductive: Find patterns from examples and generalize (best for sequences, pattern recognition)
Abductive: Work backward from the desired answer to find the best explanation (best for constraint satisfaction, puzzles)
Enumerate: Systematically list and check all possibilities (best for combinatorics, small search spaces)

The algorithm proceeds in five steps: (1) Propose strategies ranked by applicability, (2) Execute top-k strategies with $N$ reasoning traces each, (3) Score each strategy by agreement ratio among its traces, (4) Select the answer from the highest-scoring strategy via majority vote, and (5) optionally Reconcile if confidence is low.

We compare against three baselines: (1) Direct zero-shot prompting (1 API call), (2) Chain-of-Thought with greedy decoding (1 API call), and (3) Self-Consistency \cite{wang2023selfconsistency} with $k=5$ samples (5 API calls). Strategy Search uses approximately 14 API calls per problem (1 proposal + 4 strategies × 3 samples + meta-check).

\paragraph{Experimental Setup} We evaluate on three benchmarks: GSM8K (grade-school math), MATH (competition mathematics), and BBH (BIG-Bench Hard logical reasoning, 8 subtasks). All experiments use gpt-4o-mini with 100 examples per benchmark and a fixed seed for reproducibility.

\paragraph{Results} Table~\ref{tab:strategy_results} presents our main findings.  Strategy Search achieves the best performance on MATH (50\%), outperforming Self-Consistency by 2 percentage points. This improvement comes from problems where non-deductive strategies succeed—the enumerate strategy, when selected (10\% of cases), achieves 60\% accuracy on MATH problems. However, Strategy Search underperforms on BBH (88\% vs.\ 92\% for Self-Consistency), suggesting that for tasks well-suited to deductive reasoning, explicit strategy diversification can introduce noise.  A surprising finding is that direct prompting (93\%) outperforms Chain-of-Thought (89\%) on GSM8K, consistent with recent observations that capable models may introduce errors through verbose reasoning on simple problems.

\subsection{Model Merging}
\paragraph{Implementation} We implement MALS (Merging by Adaptive Layerwise Sparsity Allocation), a model merging algorithm that adaptively allocates sparsity across layers based on task conflict metrics. The core idea is to assign higher sparsity to high-conflict layers (where task vectors interfere) while preserving capacity in low-conflict layers. The core of MALS is a three-stage procedure: (1) compute per-layer conflict scores, (2) compute per-layer importance scores, (3) solve a constrained allocation problem that distributes sparsity non-uniformly across layers, specifically, given $T$ task vectors $\{\tau_i\}_{i=1}^{T}$, where each task vector is defined as $\tau_i = \theta_i^{\mathrm{ft}} - \theta^{\mathrm{pre}}$ (difference between parameters of fine-tuned and pre-trained model), MALS allocates a distinct sparsity level $s_l$ to each layer $l$ by jointly considering inter-task conflict and layer importance. The allocation proceeds in three stages.

\paragraph{Layerwise Conflict Scoring}
For each layer $l$ and each pair of tasks $(i, j)$, we measure conflict through two complementary signals. The first is the absolute Pearson correlation of the flattened weight updates:
\begin{equation}
    \rho_l^{(i,j)} = \frac{(\tau_i^l - \bar{\tau}_i^l)^\top (\tau_j^l - \bar{\tau}_j^l)}{\| \tau_i^l - \bar{\tau}_i^l \| \cdot \| \tau_j^l - \bar{\tau}_j^l \|},
\end{equation}
where $\tau_i^l$ denotes the vectorized task vector for task $i$ at layer $l$ and $\bar{\tau}_i^l$ its element-wise mean. High $|\rho_l^{(i,j)}|$ indicates that two tasks modify the same parameters in correlated (or anti-correlated) directions, both of which can cause interference during merging.

The second signal is the sign disagreement ratio, which captures the fraction of parameter positions where two task vectors pull in opposite directions:
\begin{equation}
    d_l^{(i,j)} = \frac{2 \cdot |\{k : \operatorname{sign}(\tau_{i,k}^l) \cdot \operatorname{sign}(\tau_{j,k}^l) < 0\}|}{|\{k : \tau_{i,k}^l \neq 0\}| + |\{k : \tau_{j,k}^l \neq 0\}|}.
\end{equation}

We combine these two signals with equal weight and average over all $\binom{T}{2}$ task pairs to obtain a single conflict score per layer:
\begin{equation}
    c_l = \frac{1}{\binom{T}{2}} \sum_{i < j} \left[ \tfrac{1}{2} \left| \rho_l^{(i,j)} \right| + \tfrac{1}{2} \, d_l^{(i,j)} \right].
\end{equation}

\paragraph{Layer Importance}
To prevent over-pruning of layers that carry substantial task-specific information, we define a magnitude-based importance score as the mean absolute value of the task vector entries, averaged across tasks:
\begin{equation}
    m_l = \frac{1}{T} \sum_{i=1}^{T} \operatorname{mean}\!\left(|\tau_i^l|\right).
\end{equation}
Layers with larger fine-tuning updates are considered more important and receive lower sparsity.

\paragraph{Adaptive Allocation}
Both $c_l$ and $m_l$ are min-max normalized to $[0,1]$:
\begin{equation}
\begin{split}
    \hat{c}_l = \frac{c_l - \min_{l'} c_{l'}}{\max_{l'} c_{l'} - \min_{l'} c_{l'}}, \qquad \\
    \hat{m}_l = \frac{m_l - \min_{l'} m_{l'}}{\max_{l'} m_{l'} - \min_{l'} m_{l'}}.
\end{split}
\end{equation}
A raw allocation score for each layer trades off conflict reduction against importance preservation with hyperparameters $\alpha$ and $\beta$:
\begin{equation}
    r_l = \alpha \, \hat{c}_l - \beta \, \hat{m}_l.
\end{equation}
A softmax transformation converts these scores into a non-negative distribution over layers:
\begin{equation}
    w_l = \frac{\exp(r_l - \max_{l'} r_{l'})}{\sum_{l'} \exp(r_{l'} - \max_{l'} r_{l'})},
\end{equation}
and the initial per-layer sparsity is obtained by linearly mapping $w_l$ into the allowed range $[s_{\min}, s_{\max}]$:
\begin{equation}
    s_l^{(0)} = s_{\min} + w_l \cdot (s_{\max} - s_{\min}).
\end{equation}

To enforce the global budget constraint $\frac{1}{L}\sum_{l} s_l = s_{\mathrm{target}}$, we apply an iterative projection that alternates between an affine correction and box clipping. At each iteration, we compute a uniform shift $\delta = s_{\mathrm{target}} - \frac{1}{L}\sum_l s_l$ and update $s_l \leftarrow \operatorname{clip}(s_l + \delta,\, s_{\min},\, s_{\max})$. When clipping saturates some layers at a bound, the residual budget is redistributed among the free (non-saturated) layers $\mathcal{F} = \{l : s_{\min} < s_l < s_{\max}\}$ via a scaled correction:
\begin{equation}
\begin{split}
    \delta' = \frac{(s_{\mathrm{target}} - \bar{s}) \cdot L}{|\mathcal{F}|}, \qquad \\
    s_l \leftarrow \operatorname{clip}(s_l + \delta',\, s_{\min},\, s_{\max}), \quad \forall\, l \in \mathcal{F}.
\end{split}
\end{equation}
This procedure converges when $|\frac{1}{L}\sum_l s_l - s_{\mathrm{target}}| < \epsilon$, which in practice requires fewer than ten iterations.

After allocation, each task vector is sparsified by retaining only the top-$(1 - s_l)$ fraction of entries by magnitude at each layer, yielding layerwise masks that concentrate surviving parameters in low-conflict, high-importance regions of the network. The sparsified task vectors then proceed to sign election and disjoint merging, producing the final merged model as $\theta^{\mathrm{merged}} = \theta^{\mathrm{pre}} + \lambda \, \tau^{\mathrm{merged}}$.

\paragraph{Experimental Setup} For language model experiments, we use Mistral-7B~\citep{jiang2023mistral7b} as the base model and merge three fine-tuned variants: WizardMath-7B-V1.1 (math reasoning), OpenHermes-2.5-Mistral-7B (general reasoning), and zephyr-7b-beta (chat/QA). We evaluate on GSM8K, ARC-Easy, ARC-Challenge, HellaSwag, and perplexity (WikiText-2) with 200 samples per benchmark (randomly sampled). For vision experiments, we use ViT-B/16 pretrained on ImageNet and fine-tune four task-specific models on CIFAR-100 subsets (animals, vehicles, objects, nature), evaluating on the full CIFAR-100 test set. All experiments use 50\% target sparsity.

We compare MALS against several baselines: (1) Simple Averaging, which uniformly averages all task vectors; (2) Uniform Sparsity, which applies identical magnitude-based pruning across all layers; and (3) TIES-Merging \cite{yadav2023tiesmerging}, which combines trimming, sign election, and disjoint merging.

\paragraph{Results} Table~\ref{tab:merging_results} presents our main results. A key finding is that \textbf{sign election is domain-dependent}: while TIES-Merging achieves the best vision accuracy (52.95\%), it catastrophically fails on LLM reasoning tasks (12.5\% ARC-Easy vs.~75.5\% for MALS without sign election). We attribute this to the diversity of LLM fine-tuning objectives—math, reasoning, and chat tasks induce conflicting parameter updates, causing TIES's sign election to zero out critical weights.

\begin{table*}[t]
\centering
\caption{Model merging results on LLM (Mistral-7B) and Vision (ViT-B/16) tasks. Best results per column in \textbf{bold}. $\downarrow$ indicates lower is better.}
\label{tab:merging_results}
\resizebox{0.9\textwidth}{!}{
\begin{tabular}{l|ccccc|cc}
\toprule
\multirow{2}{*}{\textbf{Method}} 
& \multicolumn{5}{c|}{\textbf{LLM (Mistral-7B)}} 
& \multicolumn{2}{c}{\textbf{Vision (ViT-B/16)}} \\
\cmidrule(lr){2-6} \cmidrule(lr){7-8}
& GSM8K & ARC-E & ARC-C & HSwag & PPL $\downarrow$ & Acc. & F1 \\
\midrule
Simple Average & 0.530 & 0.430 & 0.290 & 0.620 & 13.06 & 0.473 & 0.453 \\
Uniform Sparsity & 0.480 & 0.750 & 0.605 & 0.605 & 12.21 & 0.425 & 0.399 \\
TIES-Merging & 0.545 & 0.125 & 0.090 & \textbf{0.655} & 14.57 & \textbf{0.530} & \textbf{0.515} \\
TIES (w/ sign) & 0.545 & 0.455 & 0.310 & 0.630 & 12.97 & -- & -- \\
\midrule
\textbf{MALS (ours)} & 0.525 & \textbf{0.755} & \textbf{0.605} & 0.625 & \textbf{12.22} & 0.493 & 0.472 \\
\textbf{MALS (w/ sign)} & \textbf{0.560} & 0.475 & 0.315 & 0.625 & 12.61 & -- & -- \\

\bottomrule
\end{tabular}
}
\end{table*}

\begin{table}[t]
\centering
\caption{Accuracy of the proposed Strategy Search method on three reasoning benchmarks. \textbf{Bold} means the best performance.
}
\resizebox{0.98\linewidth}{!}{
\begin{tabular}{lccc}
\toprule[1.5pt]
\textbf{Method} & \textbf{GSM8K} & \textbf{MATH} & \textbf{BBH} \\
\midrule[1pt]
Direct  & 93.0 &  46.0 & 84.0  \\
CoT & 89.0 & 42.0 & 90.0 \\
Self-Consistency & 95.0 & 48.0 & \textbf{92.0} \\
Strategy Search (ours) & \textbf{95.0} & \textbf{50.0} & 88.0\\
\bottomrule[1.5pt]
\end{tabular}
}
\label{tab:strategy_results}
\end{table}



\section{FAS Robustness Validation}
\label{sec:fas_robustness}

\subsection{Statistical Significance of FAS Improvements}
\label{sec:fas_significance}

To quantify the uncertainty of the main FAS comparisons, we conduct paired significance tests on the logged per-instance scores for all 819 evaluation instances. Each comparison is paired by input, so the two systems in a pair are evaluated on the same research question and inspiring papers. We report a paired bootstrap with 10{,}000 resamples and the Wilcoxon signed-rank test. Table~\ref{tab:fas_significance} reports differences on the same overall-FAS scale as Table~\ref{tab:main_results}.

\begin{table*}[t]
\centering
\small
\setlength{\tabcolsep}{6pt}
\caption{Paired significance tests for overall FAS on all 819 evaluation instances. The bootstrap uses 10{,}000 paired resamples. All confidence intervals exclude zero, and both tests reject the null hypothesis for every comparison shown.}
\label{tab:fas_significance}
\resizebox{0.88\textwidth}{!}{%
\begin{tabular}{lcccc}
\toprule
\textbf{Comparison} & \textbf{Mean diff.} & \textbf{95\% bootstrap CI} & \textbf{Bootstrap $p$} & \textbf{Wilcoxon $p$} \\
\midrule
Llama-3.1-8B: Ours vs. prompting & $+3.17$ & $[2.31,\,4.04]$ & $<10^{-4}$ & $1.54\times10^{-13}$ \\
Qwen2.5-7B: Ours vs. prompting & $+2.89$ & $[2.25,\,3.55]$ & $<10^{-4}$ & $3.35\times10^{-17}$ \\
Qwen2.5-14B: Ours vs. prompting & $+6.69$ & $[6.13,\,7.25]$ & $<10^{-4}$ & $5.10\times10^{-82}$ \\
Qwen2.5-14B: Ours vs. w/o stepwise & $+3.60$ & $[3.06,\,4.14]$ & $<10^{-4}$ & $3.84\times10^{-34}$ \\
\bottomrule
\end{tabular}%
}
\end{table*}

At the component level, \emph{Hypothesis}, \emph{Proposed Method}, and \emph{Novelty Claims} each improve significantly over prompting on all three backbones (Wilcoxon $p<10^{-11}$). \emph{Experimental Details} is the exception: the change is not significant on either Qwen backbone and is significantly lower for Llama-3.1-8B, consistent with Table~\ref{tab:main_results} (51.9 vs. 55.0) and our observation that experimental planning remains the most challenging dimension.

All proposal generators use the same inference temperature of 0.7, while the FAS semantic judge uses temperature 0.1 (Appendix~\ref{sec:semantic_judge}) and the multi-dimensional judge uses temperature 0 (Appendix~\ref{sec:ana_llm_judge}). Pairing controls for between-instance variation by comparing systems on identical inputs; the strong effect direction also replicates independently across three model backbones, expert human evaluation (Section~\ref{sec:human_eval}), and multi-dimensional judging (Section~\ref{subsec:llm_judge}). Repeated decoding seeds would be required to separately estimate seed-to-seed generation variance, which is not measured by these instance-level tests.

\begin{table}[t]
\centering
\small
\setlength{\tabcolsep}{4pt}
\caption{Robustness of FAS under evaluation-pipeline variations (n=300). We vary retrieval depth ($k$), embedding model, and judge model from the default setting (shaded). Rankings remain consistent (\textsc{Ours} $>$ Untuned $>$ CoI). Avg.\ Pearson ($r$) and Spearman ($\rho$) are instance-level correlations with the baseline.}
\label{tab:robustness}
\resizebox{0.98\linewidth}{!}{
\begin{tabular}{ccc ccc cc}
\toprule
\textbf{$k$} & \textbf{Embed} & \textbf{Judge} & \textbf{\textsc{Ours}} & \textbf{Untuned} & \textbf{CoI} & \textbf{$r$} & \textbf{$\rho$} \\
\midrule
\rowcolor{lightgray}
10 & 3-large & 4.1-mini & 6.94 & 6.46 & 6.07 & -- & -- \\
5  & 3-large & 4.1-mini & 6.93 & 6.46 & 6.05 & 0.946 & 0.931 \\
10 & 3-small & 4.1-mini & 6.74 & 6.36 & 5.93 & 0.712 & 0.771 \\
10 & 3-large & 4o-mini  & 6.72 & 6.57 & 6.25 & 0.705 & 0.664 \\
\bottomrule
\end{tabular}
}
\end{table}

To verify the robustness of our evaluation framework, we vary three components of FAS independently: retrieval depth, embedding model, and judge model. Starting from the default configuration ($k{=}10$, \texttt{text-embedding-3-large}, GPT-4.1-mini), we re-evaluate 300 randomly sampled test instances under each variant setting.

As shown in Table~\ref{tab:robustness}, the model ordering remains consistent across all configurations (\textsc{Ours} $>$ Untuned $>$ CoI). Reducing retrieval depth from $k{=}10$ to $k{=}5$ yields nearly identical absolute scores and very high instance-level correlation with the baseline (Avg.\ $r{=}0.946$, $\rho{=}0.931$), indicating that the top candidates typically appear within the top-5 retrieval results. Using a smaller embedding model (\texttt{text-embedding-3-small}) lowers absolute scores but preserves relative comparisons (Avg.\ $r{=}0.712$, $\rho{=}0.771$); we attribute the reduced correlation primarily to differences in similarity calibration and the smaller embedding model being less sensitive to fine-grained semantic distinctions. Replacing the judge with GPT-4o-mini similarly maintains the same ordering (Avg.\ $r{=}0.705$, $\rho{=}0.664$), consistent with weaker discrimination of subtle proposal--paper differences and modest shifts in score calibration. Overall, while absolute FAS values vary with embedding/judge capacity, the qualitative conclusions and model rankings remain stable under reasonable pipeline variations.

\paragraph{Inter-Judge Agreement}
To further validate that FAS results are not artifacts of judge capability, we re-evaluate test proposals using two additional judges spanning two capability tiers: GPT-4o-mini (same tier as our primary judge) and GPT-5-mini (substantially more capable), under the identical retrieval configuration (top-$k{=}10$, \texttt{text-embedding-3-large}). Per-example agreement statistics with our primary judge (GPT-4.1-mini) are reported in Table~\ref{tab:inter_judge}.

\begin{table}[t]
\centering
\small
\setlength{\tabcolsep}{4pt}
\caption{Inter-judge agreement with the primary judge (GPT-4.1-mini). Both judges yield the identical model ordering (\textsc{Ours} $>$ CoI $>$ Untuned).}
\label{tab:inter_judge}
\resizebox{0.98\linewidth}{!}{
\begin{tabular}{llccc}
\toprule
\textbf{Judge} & \textbf{Model} & \textbf{Pearson $r$} & \textbf{Spearman $\rho$} & \textbf{Within 1 pt} \\
\midrule
\multirow{3}{*}{GPT-4o-mini}
& Stepwise CoT & 0.681 & 0.614 & 98.7\% \\
& Untuned      & 0.694 & 0.684 & 97.7\% \\
& CoI          & 0.740 & 0.693 & 94.3\% \\
\midrule
\multirow{3}{*}{GPT-5-mini}
& Stepwise CoT & 0.803 & 0.796 & 86.0\% \\
& Untuned      & 0.796 & 0.781 & 88.6\% \\
& CoI          & 0.824 & 0.765 & 89.0\% \\
\bottomrule
\end{tabular}
}
\end{table}

Both judges show strong per-example correlation with GPT-4.1-mini and rate 86\%--99\% of proposals within 1 point on the 1--10 scale. Crucially, both yield the identical model ordering (Stepwise CoT $>$ Chain-of-Ideas $>$ untuned baseline) as our primary judge. The fact that a substantially more capable judge (GPT-5-mini) recovers the same ranking with strong per-example correlation indicates that the reported improvements are not artifacts of judge capability.

\section{Qualitative Analysis}
\label{sec:qualitative}
To verify that higher evaluation scores correspond to genuine alignment with human research rather than superficial score inflation, we conduct two case studies comparing proposals from our stepwise-CoT model, the untuned baseline, and Chain-of-Ideas against human-derived proposals reconstructed from published papers. The results are shown in Table~\ref{tab:qual_case1} and Table~\ref{tab:qual_case2}.

\paragraph{Case 1: TypedThinker (LLM reasoning)}
The published paper proposes a framework with three tightly integrated components: a meta-thinker that predicts reasoning type effectiveness, an explicit demonstration memory for type-specific few-shot retrieval, and a fine-tuned reasoner with weighted voting across reasoning types.
Our model generates a proposal that captures the same core insight---explicit strategy selection and coordination across diverse reasoning types (deductive, inductive, abductive, analogical)---and designs a concrete mechanism for it (strategy biasing with modular operations), receiving scores of 7/10 on both hypothesis and method.
In contrast, the untuned model proposes a diffuse collection of loosely connected components---graph representations, multiagent debate, and world model integration---none of which appear in the published paper (method: 3/10).
CoI similarly defaults to a generic ``hybrid framework'' with multi-agent debate, lacking the specific meta-thinker or demonstration memory design (method: 4/10).
Notably, while all three models reproduce a similar research question (since it is provided in the input), the quality divergence emerges in \emph{how} they propose to solve it: our model converges on a focused, multi-component architecture that mirrors the human design, whereas the baselines resort to enumerating popular techniques without a coherent design rationale.

\paragraph{Case 2: Retrieval Head (long-context factuality)}
The published paper identifies sparse, specialized attention heads---termed ``retrieval heads''---that perform copy-paste retrieval from long contexts, validated through Needle-in-a-Haystack tests across multiple model families.
Our model proposes ``Long-Context Retrieval Heads,'' nearly matching the published title, and correctly hypothesizes that a subset of attention heads act as retrieval mechanisms that are causally responsible for factuality (hypothesis: 9/10).
It further proposes a concrete identification method via PCA-based analysis and causal validation through ablation (method: 7/10).
The published paper was even retrieved as the most similar paper in the embedding-based evaluation, confirming semantic alignment beyond surface-level wording.
The untuned baseline, while identifying the correct general direction, dilutes its proposal across three loosely connected tracks---mechanistic analysis, chain-of-thought evaluation, and KV cache compression---without converging on the focused retrieval head mechanism (method: 5/10).
CoI frames the problem around ``induction heads'' rather than the more precise ``retrieval heads'' concept and proposes modifying attention architectures, diverging from the interpretability-focused approach of the actual paper (method: 5/10).

These case studies illustrate a consistent pattern: our stepwise reasoning procedure enables the model to synthesize insights from the input literature into a \emph{focused, methodologically specific} proposal that closely mirrors what human researchers ultimately published, whereas the baselines tend to produce generic frameworks that enumerate popular techniques without a coherent research narrative.

\begin{table*}[t]
\centering
\footnotesize
\caption{Case Study 1---\emph{TypedThinker: Typed Thinking Improves Large Language Model Reasoning}. The human-derived proposal designs a meta-thinker that selects among reasoning types with demonstration memory and weighted voting. Our model captures this core design, while the baselines propose generic frameworks with unrelated components.}
\label{tab:qual_case1}
\begin{tabularx}{\linewidth}{p{0.55cm} p{0.85cm} p{4cm} X}
\toprule
\textbf{Dim.} & \textbf{Model} & \textbf{Human-Derived Reference} & \textbf{Generated Proposal} \\
\midrule

\multirow{3}{*}{\rotatebox[origin=c]{90}{\normalsize Hypothesis}}
  & \makecell[l]{\textsc{Ours}\\\scriptsize(7/10)}
  & \multirow{3}{=}{\raggedright Incorporating \textbf{explicit selection and demonstration} of diverse reasoning types (deductive, inductive, abductive, analogical) for each problem instance will enable LLMs to solve a broader range of problems more effectively than approaches relying on a single or undifferentiated reasoning strategy.}
  & \textbf{Explicitly biasing and coordinating} the use of high-level reasoning strategies within LLMs, particularly by introducing \textbf{strategy biasing and modular reasoning operations}, will significantly enhance their performance on complex logical and mathematical tasks compared to standard prompting methods. \\
\cmidrule(l){2-2} \cmidrule(l){4-4}
  & \makecell[l]{Untuned\\\scriptsize(4/10)}
  &
  & Integrating diverse reasoning strategies within a unified framework will significantly enhance the problem-solving capabilities of large language models, particularly in complex logical and mathematical tasks, compared to models relying solely on deductive reasoning. \\
\cmidrule(l){2-2} \cmidrule(l){4-4}
  & \makecell[l]{CoI\\\scriptsize(4/10)}
  &
  & By integrating a Hybrid Reasoning Framework that includes abductive, inductive, deductive, and analogical reasoning modalities, along with \textbf{dynamic strategy selection and enhanced interactive mechanisms}, LLMs can significantly enhance their problem-solving capabilities, particularly in tasks involving incomplete data, pattern recognition, and prediction. \\

\midrule

\multirow{3}{*}{\rotatebox[origin=c]{90}{\normalsize Method}}
  & \makecell[l]{\textsc{Ours}\\\scriptsize(7/10)}
  & \multirow{3}{=}{\raggedright Three main components: (1) a \textbf{meta-thinker}, fine-tuned to predict the effectiveness of each reasoning type based on empirical success rates; (2) an \textbf{explicit demonstration collection}, storing successful solutions for each type, enabling \textbf{retrieval of relevant few-shot examples}; (3) a \textbf{reasoner} applying the selected type, guided by retrieved demonstrations. The system aggregates answers via \textbf{weighted voting}.}
  & Two key components: \textbf{Strategy Biasing}---explicitly biases the LLM toward specific strategies (inductive, abductive, analogical) by conditioning the prompt on the desired strategy; \textbf{Modular Reasoning Operations}---task-adaptive reasoning functions that can be flexibly combined and executed in parallel. A \textbf{Strategy Coordination} mechanism schedules and combines the outputs of different reasoning strategies to optimize task performance, leveraging the strengths of each strategy for a given problem. \\
\cmidrule(l){2-2} \cmidrule(l){4-4}
  & \makecell[l]{Untuned\\\scriptsize(3/10)}
  &
  & \textbf{Reasoning Strategy Module}: dynamically switch between abductive, inductive, and analogical reasoning \ldots\ \textbf{Graph Representation}: graph-based structure where nodes represent reasoning steps and edges represent dependencies. \textbf{Multiagent Debate}: multiple LLM instances generate and critique reasoning paths using different strategies. \textbf{World Model Integration}: incorporate a world model to simulate state transitions and integrate planning algorithms to guide the reasoning process. \\
\cmidrule(l){2-2} \cmidrule(l){4-4}
  & \makecell[l]{CoI\\\scriptsize(4/10)}
  &
  & Three core components: Integrated Reasoning Modalities, Dynamic Strategy Selection, and Enhanced Interactive Mechanisms. Each reasoning type is a distinct module, designed to operate independently yet collaboratively. A decision-making layer evaluates task requirements and selects the most suitable strategy. Leveraging the \textbf{multiagent debate framework}, the method facilitates collaboration among multiple model instances, refining and validating reasoning outcomes through iterative feedback. \\

\midrule

\multirow{3}{*}{\rotatebox[origin=c]{90}{\normalsize Experiment}}
  & \makecell[l]{\textsc{Ours}\\\scriptsize(6/10)}
  & \multirow{3}{=}{\raggedright Open-source LLMs (\textbf{Mistral 7B, LLaMA3 8B, Qwen2-7B}) on logical reasoning (\textbf{LogiQA, BBH}) and math (\textbf{GSM8K, MATH}). Baselines: few-shot prompting, CoT selection, self-discover, mixture-of-reasoning. Generalization to unseen domains; transferability to GPT-4o. \textbf{Ablation studies} on each component.}
  & Benchmarks including \textbf{GSM8K, AIME, MATH}, HumanEval-Analogies, AIME-Logic, and MATH-Logic. Exact-match accuracy with analysis of task-specific correctness and robustness to prompt imperfections. Baselines: standard CoT, CoT with Self-Consistency, GoT, and Debate. \textbf{Ablation studies} for Strategy Biasing, Modular Reasoning Operations, and Strategy Coordination. Efficiency and robustness to model scale evaluated. \\
\cmidrule(l){2-2} \cmidrule(l){4-4}
  & \makecell[l]{Untuned\\\scriptsize(4/10)}
  &
  & Collect tasks from benchmarks like \textbf{BIG-Bench Hard}. Train LLM using standard pre-training and fine-tuning. Test on curated tasks. Metrics: accuracy, consistency, and creativity. Compare performance with and without the proposed framework. Identify tasks where the model shows emergent reasoning capabilities. \\
\cmidrule(l){2-2} \cmidrule(l){4-4}
  & \makecell[l]{CoI\\\scriptsize(3/10)}
  &
  & Datasets: BIG-Bench Suite, MATH, Winogrande, COPA. Baselines: standard LLMs, self-consistency, graph-based reasoning, multi-agent debate, world model integrated models. Metrics: accuracy, latency, consistency, adaptability score. Fine-tuning on respective datasets, deploying on test sets, and conducting qualitative analysis. \\

\bottomrule
\end{tabularx}
\end{table*}

\begin{table*}[t]
\centering
\footnotesize
\caption{Case Study 2---\emph{Retrieval Head Mechanistically Explains Long-Context Factuality}. The human-derived proposal identifies sparse ``retrieval heads'' responsible for copy-paste retrieval, validated via Needle-in-a-Haystack tests. Our model nearly matches this design, correctly naming the concept and proposing causal ablation. The baselines dilute across loosely connected tracks (Untuned) or misidentify the mechanism as ``induction heads'' (CoI).}
\label{tab:qual_case2}
\begin{tabularx}{\linewidth}{p{0.55cm} p{0.85cm} p{4cm} X}
\toprule
\textbf{Dim.} & \textbf{Model} & \textbf{Human-Derived Reference} & \textbf{Generated Proposal} \\
\midrule

\multirow{3}{*}{\rotatebox[origin=c]{90}{\normalsize Hypothesis}}
  & \makecell[l]{\textsc{Ours}\\\scriptsize(9/10)}
  & \multirow{3}{=}{\raggedright A small, specialized subset of attention heads, termed \textbf{retrieval heads}, are primarily responsible for retrieval from long contexts. These are \textbf{universal, sparse, intrinsic} to pretrained models, dynamically activated, and \textbf{causally linked} to factuality and complex reasoning.}
  & A subset of attention heads in long-context LLMs act as \textbf{retrieval heads}, selectively \textbf{copying and retrieving} information from arbitrary positions in the input sequence, which is \textbf{causally responsible} for improved factuality and reasoning performance. \\
\cmidrule(l){2-2} \cmidrule(l){4-4}
  & \makecell[l]{Untuned\\\scriptsize(6/10)}
  &
  & Specific attention head circuits in LLMs are responsible for long-context retrieval, and these mechanisms significantly enhance the model's factuality and reasoning abilities in downstream tasks. \\
\cmidrule(l){2-2} \cmidrule(l){4-4}
  & \makecell[l]{CoI\\\scriptsize(6/10)}
  &
  & It is hypothesized that \textbf{induction heads} significantly contribute to long-context retrieval through specific interaction patterns with other attention heads, thereby enhancing the model's ability to maintain and leverage long-term dependencies. Furthermore \ldots\ novel prompting strategies and architectural designs can improve long-context retrieval and reasoning. \\

\midrule

\multirow{3}{*}{\rotatebox[origin=c]{90}{\normalsize Method}}
  & \makecell[l]{\textsc{Ours}\\\scriptsize(7/10)}
  & \multirow{3}{=}{\raggedright Define a \textbf{retrieval score} for each attention head, quantifying \textbf{copy-paste behavior} during autoregressive decoding. \textbf{Needle-in-a-Haystack} tests with unique QA pairs embedded at random positions \ldots\ Retrieval scores computed across diverse contexts and model variants. Examine \textbf{universality, sparsity, intrinsic nature}, and dynamic activation across model families, scales, and fine-tuning types.}
  & Identify and characterize \textbf{retrieval heads}---attention heads that selectively copy and retrieve from arbitrary positions. Cluster heads based on copying behavior using \textbf{PCA of per-token loss vectors}; retrieval heads identified as those exhibiting long-range copying across multiple training snapshots. Causal role validated through \textbf{ablation studies}, where retrieval heads are removed or replaced and impact on retrieval and downstream reasoning is measured. Retention patterns in the \textbf{KV cache} analyzed for memory usage and efficiency. \\
\cmidrule(l){2-2} \cmidrule(l){4-4}
  & \makecell[l]{Untuned\\\scriptsize(5/10)}
  &
  & Three separate tracks: (1) \textbf{Mechanistic Analysis}---per-token loss PCA, identify attention heads via sequence copying tasks, architectural perturbations and direct ablations; (2) \textbf{Chain-of-Thought Prompting}---evaluate reasoning with and without identified retrieval heads, compare to standard prompting; (3) \textbf{KV Cache Compression}---adaptive techniques (e.g., FastGen), use retrieval heads to inform compression policies, ensure critical context retained. \\
\cmidrule(l){2-2} \cmidrule(l){4-4}
  & \makecell[l]{CoI\\\scriptsize(5/10)}
  &
  & Multi-faceted approach: mechanistic analysis of attention mechanisms, focusing on \textbf{induction heads} and their interactions \ldots\ New prompting strategies based on chain-of-thought prompting to encourage engagement with long-range dependencies. Architectural designs to prioritize long-term context, including modifications to the attention mechanism and novel training objectives. \\

\midrule

\multirow{3}{*}{\rotatebox[origin=c]{90}{\normalsize Experiment}}
  & \makecell[l]{\textsc{Ours}\\\scriptsize(6/10)}
  & \multirow{3}{=}{\raggedright LLMs including \textbf{Llama-2, Yi, Qwen, Mistral, Mixtral}, scales 6B--34B, regimes: base, chat, MoE, SFT, RLHF. Primary: \textbf{Needle-in-a-Haystack} with ${\sim}$600 instances per model, context 1K--50K tokens. Downstream: extractive QA, CoT reasoning (\textbf{MMLU, MuSiQue, GSM8K}). \textbf{Masking} retrieval vs.\ random heads.}
  & Suite of long-context LLMs with and without attention sinks. Primary: long-context retrieval tasks prompting models to retrieve from arbitrary positions in long sequences. Downstream: mathematical reasoning (\textbf{GSM8K, AIME, MATH}). Metrics: retrieval accuracy (top-$k$), reasoning accuracy/F1, factuality scores. \textbf{Ablation studies} systematically remove or replace retrieval heads to quantify causal impact. Synthetic datasets with long-range dependencies validate copying behavior. \\
\cmidrule(l){2-2} \cmidrule(l){4-4}
  & \makecell[l]{Untuned\\\scriptsize(5/10)}
  &
  & Train multiple LLMs (\textbf{GPT-3, T5, BERT}) with varying architectures. Per-token loss PCA to identify retrieval heads. Chain-of-thought prompts for reasoning tasks (arithmetic, commonsense, symbolic). Adaptive KV cache compression (FastGen). Metrics: accuracy, F1, perplexity. Compare to fixed-policy baselines to demonstrate effectiveness. \\
\cmidrule(l){2-2} \cmidrule(l){4-4}
  & \makecell[l]{CoI\\\scriptsize(5/10)}
  &
  & Benchmarks: \textbf{GLUE, SuperGLUE, Long Range Arena}, supplemented by custom longer-context datasets. Baselines: BERT, RoBERTa, T5, with induction heads and adaptive KV cache compression. Metrics: factuality, coherence, downstream performance, memory efficiency. Cross-validation and ablation studies to validate findings. \\

\bottomrule
\end{tabularx}
\end{table*}

\section{Implementation Details}
\label{sec:exp_details}

\subsection{Baselines}
\textbf{AI-Reseacher}: For controlled comparison, we remove the retrieval stage and provide the same inspiring papers to all methods. 
The model first generates a structured seed idea (problem, motivation, method, and experiment plan), which is then expanded into a proposal and formatted into our evaluation structure.

\textbf{Chain-of-Ideas}: In our setting, we treat the provided inspiring papers as the idea chain and prompt the model to analyze their evolution and generate a proposal aligned with the predicted research direction. 
The generated output is then formatted to match our proposal structure.

\begin{table}[t]
\centering
\small
\setlength{\tabcolsep}{4pt}
\caption{SFT training configuration for all the fine-tuning. Effective batch size equals batch size $\times$ gradient accumulation steps.}
\label{tab:train_details}
\resizebox{0.8\linewidth}{!}{
\begin{tabular}{l c}
\toprule
\textbf{Setting} & \textbf{Value} \\
\midrule
Epochs & 2 \\
Max sequence length & 8000 \\
Batch size (per device) & 1 \\
Gradient accumulation & 4 \\
Effective batch size & 4 \\
Learning rate & $2\times10^{-5}$ \\
Precision & bf16 \\
LoRA rank $r$ & 16 \\
LoRA $\alpha$ & 32 \\
\bottomrule
\end{tabular}
}
\end{table}

\subsection{Prompt for Proposal Generation}
\label{sec:generation_prompts}
We use different prompt variants to implement prompting-only, direct-CoT, and stepwise-CoT baselines under different input conditions. The system prompts are provided in Figure~\ref{fig:generation_system_prompts}. Additional user-side instructions used for CoT-based proposal generation are provided in Figure~\ref{fig:generation_cot_instructions}.

\begin{figure*}[t]
\centering
\begin{tcolorbox}[width=\textwidth,colback=gray!5,colframe=gray!60,boxrule=0.5pt,arc=2pt,left=4pt,right=4pt,top=4pt,bottom=4pt]
\footnotesize
\textbf{System Prompts for Proposal Generation}

\medskip
\textbf{Standard Generation Prompt}
\begin{verbatim}
You are an expert AI research scientist. Given inspiring research papers and a target research question, 
propose a novel research idea that addresses the question.

Your response should include:
- A proposed research idea with title, research question, hypothesis, proposed method, novelty claims, 
and experiment details

Format your response starting with "## Proposed Research".
\end{verbatim}

\medskip
\textbf{Research-Question-Only Prompt}
\begin{verbatim}
You are an expert AI research scientist. Given a research question, propose a novel research idea that 
addresses the question.

Your response should include:
- A proposed research idea with title, hypothesis, proposed method, novelty claims, and experiment details

Format your response starting with "## Proposed Research".
\end{verbatim}

\medskip
\textbf{Inspiring-Papers-Only Prompt}
\begin{verbatim}
You are an expert AI research scientist. Given a set of inspiring research papers, identify gaps and 
opportunities, then propose a novel research idea that builds upon their contributions.

Your response should include:
- A proposed research idea with title, research question, hypothesis, proposed method, novelty claims, 
and experiment details

Format your response starting with "## Proposed Research".
\end{verbatim}

\medskip
\textbf{Direct CoT Prompt}
\begin{verbatim}
You are an expert AI research scientist. Given inspiring research papers and a target research question, 
first reason about gaps and opportunities, then propose a novel research idea.

Your response should include:
1. A reasoning section analyzing gaps, borrowing inspiration, and synthesizing ideas
2. A proposed research idea with title, research question, hypothesis, proposed method, novelty claims, 
and experiment details

Format the reasoning with "### Gap Analysis", "### Inspiration Borrowing", "### Synthesis" sections, 
then the proposal starting with "## Proposed Research".
\end{verbatim}

\medskip
\textbf{Stepwise CoT Prompt}
\begin{verbatim}
You are an expert AI research scientist. Given inspiring research papers and a target research question,
develop a novel research idea step by step.

Your response should follow this exact structure:
1. Problem Identification reasoning (analyze gaps and inspiration)
2. Research Question + Hypothesis
3. Method Design reasoning (how to approach the problem)
4. Proposed Method + Novelty Claims
5. Experiment Design reasoning (how to validate)
6. Experiment Details

Use "### Step 1: Problem Identification", "### Step 2: Method Design Reasoning", "### Step 3: Experiment 
Design Reasoning" for reasoning sections, and "## Proposed Research" before the proposal sections.
\end{verbatim}
\end{tcolorbox}
\caption{System prompts used for different proposal-generation variants.}
\label{fig:generation_system_prompts}
\end{figure*}

\begin{figure*}[t]
\centering
\begin{tcolorbox}[width=\textwidth,colback=gray!5,colframe=gray!60,boxrule=0.5pt,arc=2pt,left=4pt,right=4pt,top=4pt,bottom=4pt]
\footnotesize
\textbf{Additional Instructions for CoT Variants}

\medskip
\textbf{Direct CoT Instruction}
\begin{verbatim}
Based on these inspiring papers, first analyze the research landscape, then propose a novel research idea.

Start by identifying gaps in the existing work, what ideas to borrow, and how to synthesize them. 
Then propose your research idea.
\end{verbatim}

\medskip
\textbf{Stepwise CoT Instruction}
\begin{verbatim}
Based on these inspiring papers, propose a novel research idea step by step.

First, analyze the gaps and identify what ideas to borrow from the inspiring papers. 
Then, formulate the research question and hypothesis. Next, reason about the method design, and propose 
the method with its novelty claims. Finally, reason about how to validate the method, and describe the 
experiment details.

Generate each section in this order:
1. reasoning
2. Research Question + Hypothesis
3. reasoning
4. Proposed Method + Novelty Claims
5. reasoning
6. Experiment Details
\end{verbatim}
\end{tcolorbox}
\caption{Additional user-side instructions used for CoT-based proposal generation.}
\label{fig:generation_cot_instructions}
\end{figure*}

\subsection{Training Details}
\label{subsec:training_details}

We fine-tune LLMs using parameter-efficient LoRA under the \texttt{accelerate} framework mainly on 4 H100 GPUs (94G). The hyperparameters are provided in Table~\ref{tab:train_details}.

\subsection{Data Details}
\label{sec:data_details}
\paragraph{Structured Proposal Representation} Each proposal $\hat{P}$ is generated in a pre-defined structured format to ensure consistency and enable fine-grained evaluation. The structure includes: \begin{itemize} \item \textbf{Research Question}: the main research question(s). \item \textbf{Hypothesis}: a concise statement of the core hypothesis. \item \textbf{Proposed Method}: a detailed description of the methodology, algorithm, or approach, including key components and technical steps. \item \textbf{Novelty Claims}: explicit statements describing intended contributions or innovations. \item \textbf{Experimental Details}: description of the experiments, including datasets, baselines, evaluation metrics, and validation protocols. \end{itemize}

To get the structured proposal, we first crawl the full papers in PDF format from OpenReview and extract the first 10 pages (since published papers on OpenReview has no more than 10 pages in main text). We then use GPT-4.1 to convert the papers into structured proposals with the prompt provided in Figure~\ref{fig:proposal_synthesis_prompt}.

\begin{figure*}[t]
\centering
\begin{tcolorbox}[width=\textwidth,colback=gray!5,colframe=gray!60,boxrule=0.5pt,arc=2pt,left=4pt,right=4pt,top=4pt,bottom=4pt]
\footnotesize
\textbf{Prompt for Proposal Target Synthesis}

\medskip
You are an expert research paper analyzer. Your task is to extract structured information from research papers and rewrite it as a \textbf{research proposal}, rather than as a summary of an existing paper.

\medskip
Extract the following information and return it in valid JSON format:

\begin{verbatim}
{
  "research_question": "The main research question(s). It should be broad and does not leak any key ideas.",
  "hypothesis": "The hypothesis or hypotheses (if any, otherwise 'Not specified')",
  "proposed_method": "A detailed description of the methodology, approach, algorithm, or method.
     Include key components, steps, and technical details. This should be 3-5 sentences minimum.",
  "experiment_details": "Description of the experiments, including datasets used, baselines compared, 
    evaluation metrics, and key experimental setup details. This should be 2-4 sentences minimum.",
  "novelty_claims": "Claims about novelty, contributions, or innovations"
}
\end{verbatim}

\medskip
\textbf{Critical Writing Style Guidelines:}
\begin{itemize}
    \item Write as a \textbf{research proposal}, not as a summary of an existing paper.
    \item Never use phrases such as ``The authors propose\ldots'', ``This paper introduces\ldots'', ``The paper presents\ldots'', ``They develop\ldots'', or ``The authors show\ldots''.
    \item Never reference the paper as an external work (e.g., ``this paper'', ``the authors'', ``they'', or ``the work'').
    \item Instead, write in proposal style: describe what \emph{is proposed}, what \emph{will be done}, or use passive voice.
    \item Good: ``The proposed method introduces a novel framework\ldots'' or ``A new approach is developed that\ldots''
    \item Good: ``The method achieves state-of-the-art results\ldots'' or ``Experiments demonstrate that\ldots''
    \item Bad: ``The authors introduce ROBUSTALPACAEVAL\ldots'' or ``This paper proposes a new benchmark\ldots''
    \item Be detailed and comprehensive, especially for \texttt{proposed\_method} and \texttt{experiment\_details}.
    \item If a field is not explicitly stated, infer from context when reasonable.
    \item If information is truly not available, use ``Not specified''.
\end{itemize}

\medskip
First give your reasoning process, then return the JSON response.

\medskip
The JSON response should be:
\begin{verbatim}
{
  "research_question": "...",
  "hypothesis": "...",
  "proposed_method": "...",
  "experiment_details": "...",
  "novelty_claims": "..."
}
\end{verbatim}
\end{tcolorbox}
\caption{Prompt used to convert a paper into a structured proposal target for supervision.}
\label{fig:proposal_synthesis_prompt}
\end{figure*}

\paragraph{Inspiring Papers Selection}
For each target paper, we retrieve all papers cited by the target paper via the Semantic Scholar API, then apply a two-stage filtering and selection process.
In the first stage, candidates are pre-filtered and ranked using a recency-weighted scoring function: papers published within two years of the target receive a full boost of 100 points, with a linear decay for papers up to five years older; papers exceeding this window receive zero recency score.
A minor citation-count tiebreaker ($\min(\log(1+c) \times 2, 20)$ points, where $c$ is the citation count) is added, and references with fewer than five citations are excluded.
The top 15 candidates by this score are retained.
In the second stage, an LLM (GPT-5-mini) analyzes the target paper's title and abstract alongside the 15 candidate references and selects the five that provided the most \emph{specific and direct} intellectual influence on the target's unique contributions.
The LLM is instructed to avoid well-known foundational works (e.g., Transformers, BERT, ResNet) and instead prefer papers that share niche methodological choices, specific problem formulations, or particular technical innovations that the target directly extends. Full prompt is provided in Figure~\ref{fig:inspiring_paper_selection_prompt}.

\begin{figure*}[t]
\centering
\begin{tcolorbox}[width=\textwidth,colback=gray!5,colframe=gray!60,boxrule=0.5pt,arc=2pt,left=4pt,right=4pt,top=4pt,bottom=4pt]
\footnotesize
\textbf{Prompt for Inspiring Paper Selection}

\medskip
You are an expert at analyzing academic paper citations to identify \textbf{specific} and \textbf{direct} intellectual influence.

\medskip
Given a paper and its references, identify which reference(s) provided the most specific and direct inspiration for the paper's unique contributions.

\medskip
\textbf{Current Paper}\\
Title: \{current\_title\}\\
Abstract: \{current\_abstract\}

\medskip
\textbf{Candidate References (papers cited by the current paper)}\\
\{candidates\_text\}

\medskip
\textbf{Task}\\
Select the \{num\_to\_select\} reference(s) that \textbf{directly} and \textbf{specifically} inspired the current paper's unique contributions.

\medskip
\textbf{Important Guidelines:}
\begin{itemize}
    \item \textbf{Avoid} selecting well-known foundational papers (e.g., Transformers, BERT, GPT, ResNet, U-Net, Chain-of-Thought) unless the current paper makes a very specific extension of that exact work.
    \item Prefer selecting papers that introduced the \textbf{specific} technique, formulation, or approach that the current paper directly builds upon.
    \item Look for papers that share niche methodological choices, specific problem formulations, or particular technical innovations.
    \item The best inspirations are papers where one can point to a \textbf{concrete} technique or idea that was directly adopted or extended.
\end{itemize}

\medskip
For each selection, provide:
\begin{enumerate}
    \item the reference number selected,
    \item the type of inspiration (\texttt{specific\_technique}, \texttt{direct\_extension}, \texttt{niche\_methodology}, \texttt{problem\_variant}, \texttt{algorithmic\_basis}),
    \item the specific ideas or techniques the current paper borrowed,
    \item a confidence score (0.0--1.0), where higher confidence indicates a more specific and direct connection,
    \item detailed reasoning explaining the specific intellectual connection.
\end{enumerate}

\medskip
\textbf{Output Format (JSON)}
\begin{verbatim}
{
  "selections": [
    {
      "reference_number": 1,
      "inspiration_type": "specific_technique",
      "key_ideas_borrowed": ["specific technique X", "particular formulation Y"],
      "confidence": 0.85,
      "reasoning": "The current paper's [specific contribution] directly extends reference 1's 
      [specific technique]..."
    }
  ]
}
\end{verbatim}

Respond with \textbf{only} valid JSON and no additional text.
\end{tcolorbox}
\caption{Prompt used to identify the most directly inspiring citations for each target paper.}
\label{fig:inspiring_paper_selection_prompt}
\end{figure*}

\paragraph{Reasoning Trace Synthesis}
\label{sec:reasoning_details}
We use GPT-5~\citep{openai_gpt5_2025} to synthesize the reasoning traces in the training data. The prompt for \textbf{CoT SFT} is provided in Figure~\ref{fig:direct_cot_prompt}, and the prompt for \textbf{Stepwise CoT SFT} is provided in Figure~\ref{fig:stepwise_cot_prompt}.

\begin{figure*}[t]
\centering
\begin{tcolorbox}[width=\textwidth,colback=gray!5,colframe=gray!60,boxrule=0.5pt,arc=2pt,left=4pt,right=4pt,top=4pt,bottom=4pt]
\footnotesize
\textbf{Prompt for Direct CoT Synthesis}

\medskip
You are an expert research scientist. Given a set of inspiring papers and a target paper, generate a reasoning process that explains how these papers could lead to new research.

\medskip
\textbf{Inspiring Papers (Direct Citations):}\\
\{inspiring\_papers\}

\medskip
\textbf{Target Paper (Research Outcome):}\\
Title: \{target\_title\}\\
Research Question: \{target\_research\_question\}\\
Hypothesis: \{target\_hypothesis\}\\
Proposed Method: \{target\_proposed\_method\}\\
Novelty Claims: \{target\_novelty\_claims\}

\medskip
\textbf{Task:}\\
Generate a structured reasoning process with exactly three sections. Write as if you are a researcher developing ideas \emph{before} creating the target paper, and do not leak the target paper's specific solutions.

\medskip
\textbf{Required Format (use these exact headers):}

\medskip
\textbf{Gap Analysis}\\
Identify 2--4 specific gaps or limitations in the inspiring papers:
\begin{itemize}
    \item Gap 1: [specific limitation]
    \item Gap 2: [what is missing]
    \item \ldots
\end{itemize}

\textbf{Inspiration Borrowing}\\
State what techniques or ideas to borrow from which papers:
\begin{itemize}
    \item From [Paper Title]: [specific technique or idea to adapt]
    \item From [Paper Title]: [framework or method to build on]
    \item \ldots
\end{itemize}

\textbf{Synthesis}\\
Explain how to combine the borrowed ideas to address the gaps (100--150 words), including:
\begin{itemize}
    \item the integration approach,
    \item key modifications needed,
    \item why this combination addresses the identified gaps.
\end{itemize}

\medskip
\textbf{Important:}
\begin{itemize}
    \item Total length: 300--500 words
    \item Be specific about paper names when borrowing ideas
    \item Do not mention ``the target paper'' or leak its solutions
\end{itemize}
\end{tcolorbox}
\caption{Prompt used to synthesize direct chain-of-thought reasoning traces from inspiring papers and a target paper outcome.}
\label{fig:direct_cot_prompt}
\end{figure*}

\begin{figure*}[t]
\centering
\begin{tcolorbox}[width=\textwidth,colback=gray!5,colframe=gray!60,boxrule=0.5pt,arc=2pt,left=4pt,right=4pt,top=4pt,bottom=4pt]
\footnotesize
\textbf{Prompt for Stepwise CoT Synthesis}

\medskip
You are an expert research scientist. Given a set of inspiring papers and a target paper, generate a step-by-step reasoning process that shows how a researcher would develop this research idea incrementally.

\medskip
\textbf{Inspiring Papers (Direct Citations):}\\
\{inspiring\_papers\}

\medskip
\textbf{Target Paper (the final research output):}\\
Title: \{target\_title\}\\
Research Question: \{target\_research\_question\}\\
Hypothesis: \{target\_hypothesis\}\\
Proposed Method: \{target\_proposed\_method\}\\
Novelty Claims: \{target\_novelty\_claims\}\\
Experiment Details: \{target\_experiment\_details\}

\medskip
\textbf{Task:}\\
Generate intermediate reasoning that bridges between proposal sections. You will produce exactly three short reasoning blocks. The final output will be interleaved as:

\medskip
\centerline{[Reasoning 1] $\rightarrow$ Research Question + Hypothesis $\rightarrow$ [Reasoning 2] $\rightarrow$ Proposed Method + Novelty Claims $\rightarrow$
[Reasoning 3] $\rightarrow$ Experiment Details}

\medskip
Write each reasoning block as described below.

\medskip
\textbf{Required Format (use these exact headers):}

\medskip
\textbf{Step 1: Problem Identification}\\
Analyze the inspiring papers to identify gaps and formulate the research direction (150--250 words):
\begin{itemize}
    \item what gaps or limitations exist in the inspiring papers,
    \item what techniques or ideas could be borrowed,
    \item how combining them suggests a specific research question.
\end{itemize}

\textbf{Step 2: Method Design Reasoning}\\
Given the research question and hypothesis, reason about how to design the method (80--120 words):
\begin{itemize}
    \item what approach would address the research question,
    \item which techniques from the inspiring papers to adapt,
    \item what makes this combination novel.
\end{itemize}

\textbf{Step 3: Experiment Design Reasoning}\\
Given the proposed method, reason about how to validate it (60--100 words):
\begin{itemize}
    \item what datasets or benchmarks are appropriate,
    \item what baselines to compare against,
    \item what metrics would demonstrate the method's effectiveness.
\end{itemize}

\medskip
\textbf{Important:}
\begin{itemize}
    \item Do not copy the target paper's sections verbatim into the reasoning.
    \item Write as forward-looking reasoning, as if you have not seen the final answer yet.
    \item Do not mention ``the target paper''; write as your own thought process.
    \item Each step should naturally lead to the next proposal section.
\end{itemize}
\end{tcolorbox}
\caption{Prompt used to synthesize stepwise chain-of-thought reasoning traces interleaved with proposal construction.}
\label{fig:stepwise_cot_prompt}
\end{figure*}

\paragraph{Statistics of Proposals}
\label{sec:proposal_statistics}
Table~\ref{tab:data_stats} summarizes the statistics of our datasets.
For training, we collect 2{,}823 examples from papers accepted at NeurIPS 2024 and ICLR 2024.
Each example pairs a prompt---comprising 5 structured inspiring papers and a research question (mean: 1{,}755 words)---with a completion containing the target proposal.
Three completion variants are generated: stepwise-CoT (909 words, including three explicit reasoning steps), CoT with gap analysis (900 words), and direct proposal without reasoning (460 words).
For evaluation, we construct a test set of 819 examples from papers at NeurIPS 2025, ICML 2025, and ICLR 2025, ensuring no temporal overlap with training data.


\paragraph{Dataset Statistics.}
Table~\ref{tab:data_stats} summarizes the statistics of our datasets.
For training, we collect 2{,}823 examples from papers accepted at NeurIPS 2024 (4{,}035 papers) and ICLR 2024 (2{,}261 papers).
Each example pairs a prompt---comprising 5 structured inspiring papers and an optional research question (mean: 1{,}755 words)---with a completion containing the target proposal.
Three completion variants are generated: stepwise-CoT (909 words, including three explicit reasoning steps), CoT with gap analysis (900 words), and direct proposal without reasoning (460 words).
For evaluation, we construct a test set of 819 examples from papers at NeurIPS 2025 (5{,}275 papers), ICML 2025 (3{,}260 papers), and ICLR 2025 (3{,}708 papers), ensuring no temporal overlap with training data.
Test prompts are shorter on average (1{,}057 words) as they include only the structured inspiring papers and research question without system instructions.

\begin{table}[!ht]
\centering
\small
\caption{Mean proposal length in words. For training completions, we report both the full output (including reasoning) and the proposal-only portion. For generated proposals, we report the proposal after stripping reasoning steps.}
\label{tab:data_stats}
\resizebox{0.95\linewidth}{!}{
\begin{tabular}{lrr}
\toprule
\textbf{Split} & \textbf{Raw} & \textbf{Proposal} \\
\midrule
\multicolumn{3}{l}{\textit{Training completions (NeurIPS'24 + ICLR'24)}} \\
\quad Stepwise-CoT & 909 & 460 \\
\quad CoT & 900 & 460 \\
\quad No-CoT & 460 & 460 \\
\midrule
\multicolumn{3}{l}{\textit{Test reference (NeurIPS'25 + ICML'25 + ICLR'25)}} \\
\quad Human-derived proposal & --- & 460 \\
\midrule
\multicolumn{3}{l}{\textit{Generated proposals}} \\
\quad Qwen-14B stepwise-CoT & 888 & 438 \\
\quad Qwen-14B CoT & 484 & 481 \\
\quad Qwen-14B no-CoT & 449 & 446 \\
\quad Qwen-14B untuned & 513 & 510 \\
\quad Qwen-7B stepwise-CoT & 892 & 436 \\
\quad Llama-8B stepwise-CoT & 932 & 366 \\
\quad CoI (Qwen-14B) & 342 & 339 \\
\quad AI-Researcher (Qwen-14B) & 351 & 348 \\
\bottomrule
\end{tabular}
}
\end{table}

\subsection{Human Evaluation}
\label{sec:human_eval_details}

\paragraph{Data Selection}
We sample 60 data points from the 35\% quality-filtered subset of our 819-example test set, stratified by research area: 42 NLP papers and 18 multimodal/vision papers (approximately 70/30).
For each data point, we construct two comparison pairs---Stepwise CoT vs.\ human-derived proposal and Stepwise CoT vs.\ prompting-only (untuned Qwen2.5-14B)---yielding 120 pairs in total.
Both sides of each pair are required to contain all five structured sections (Research Question, Hypothesis, Proposed Method, Novelty Claims, Experiment Details); pairs missing any section are excluded.
Reasoning traces are stripped from Stepwise CoT outputs before display so that annotators see only the final proposal.

\begin{figure*}
    \centering
    \includegraphics[width=0.98\linewidth]{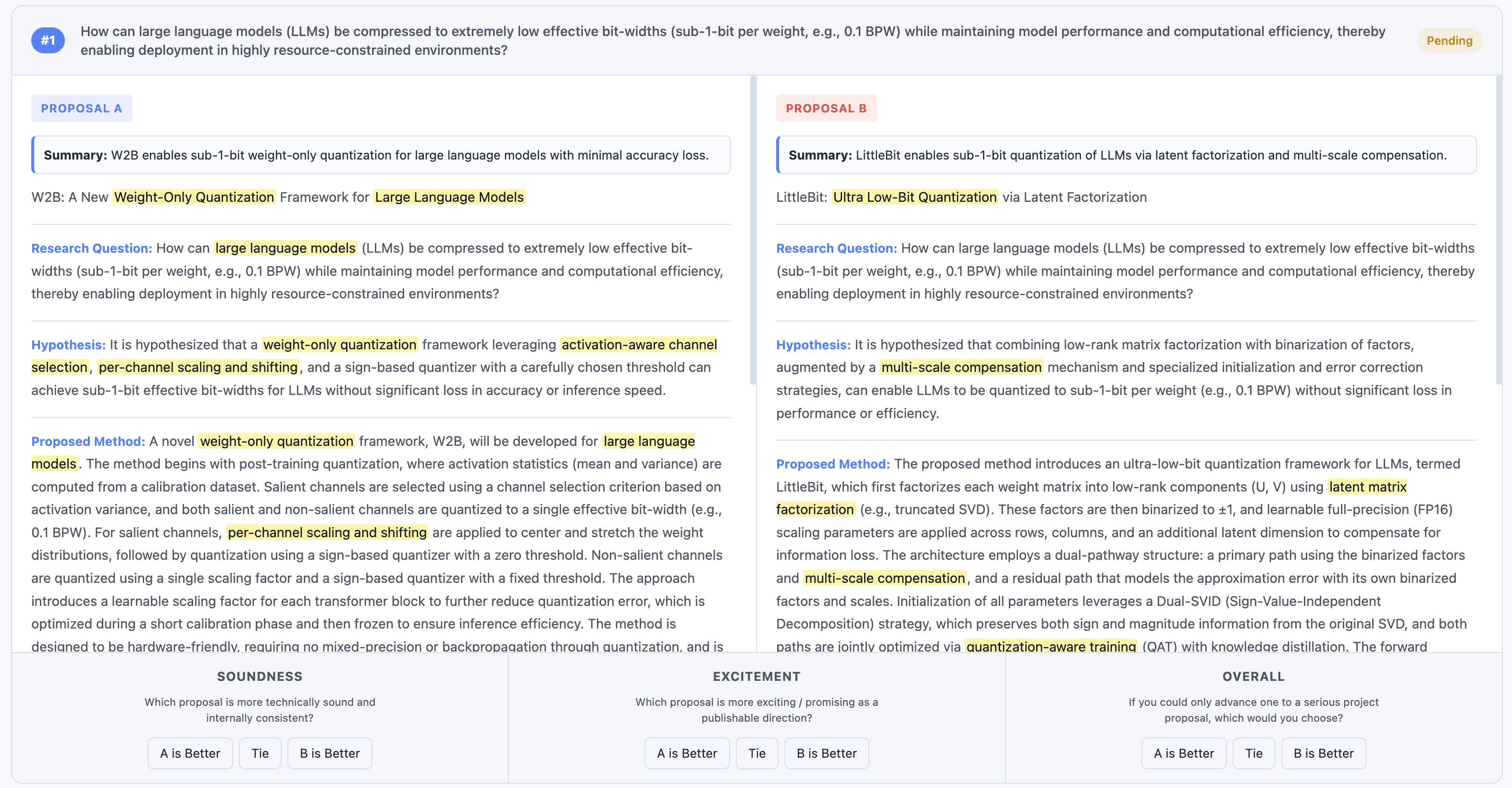}
    \caption{The annotation interface of the human evaluation.}
    \label{fig:ui}
\end{figure*}

\paragraph{Annotation Interface}
We build a web-based side-by-side comparison tool that presents each pair as ``Proposal~A'' and ``Proposal~B.'' 
We provide a screenshot of the annotation interface in Figure~\ref{fig:ui}.
To prevent positional bias, the assignment of the two proposals to sides A and B is randomized independently for each pair (with a fixed seed for reproducibility).
Annotators are not told which side is model-generated, human-derived, or from which model variant.
For each pair, annotators select one of three options---\emph{A is Better}, \emph{Tie}, or \emph{B is Better}---along each of three dimensions:
\begin{itemize}[nosep,leftmargin=1.5em]
    \item{\textbf{Soundness}: Which proposal is more technically sound and internally consistent?}
    \item \textbf{Excitement}: Which proposal is more exciting or promising as a publishable research direction?
    \item \textbf{Overall}: If you could only advance one to a serious research project, which would you choose?
\end{itemize}

\paragraph{Annotators and Batching}
The 120 pairs are divided into four batches of 30 pairs each (15 per comparison type), and each batch is independently annotated by three domain-expert graduate students with prior conference reviewing experience.
A total of 11 unique annotators participate across the four batches; no annotator sees the same pair twice.

\textbf{Aggregation.}
For each pair and dimension, we take the majority vote among the three annotators as the final judgment.
When no category receives a strict majority (e.g., one vote each for A, B, and Tie), we treat the outcome as a tie.
Win rates are computed by crediting full wins as~1 and ties as~0.5; 95\% confidence intervals are obtained via the Wilson score method.

\paragraph{Detailed Results}
Table~\ref{tab:human_eval_detail} reports the full win/tie/loss counts, win rates with 95\% confidence intervals, and the percentage of pairs where all three annotators agree (unanimity).
Against human-derived proposals, Stepwise CoT achieves near-parity across all dimensions, with win rates close to 50\% and wide confidence intervals reflecting the difficulty of the comparison.
The low unanimity (6.7--11.7\%) confirms that differences between strong model-generated and human-derived proposals are often subtle.
Against prompting-only proposals, Stepwise CoT is preferred more consistently, with higher unanimity (26.7--31.7\%) indicating clearer quality differences.

\begin{table}[!ht]
\centering
\small
\caption{Detailed human evaluation results. Win/Tie/Loss counts reflect the majority vote across three annotators per pair. Win rate treats ties as 0.5 wins. CI: Wilson score 95\% confidence interval. Unanimity: percentage of pairs where all three annotators agree.}
\label{tab:human_eval_detail}
\resizebox{1\linewidth}{!}{
\begin{tabular}{llccccc}
\toprule
\textbf{Comparison} & \textbf{Dimension} & \textbf{Win} & \textbf{Tie} & \textbf{Loss} & \textbf{Win Rate (95\% CI)} & \textbf{Unanimity} \\
\midrule
\multirow{3}{*}{\shortstack[l]{Stepwise CoT\\vs.\ Human}} 
  & Overall    & 25 & 10 & 25 & 50.0\% [37.7, 62.3] & 11.7\% \\
  & Soundness  & 21 & 11 & 28 & 44.2\% [32.3, 56.7] & 11.7\% \\
  & Excitement & 20 & 18 & 22 & 48.3\% [36.2, 60.7] & ~6.7\% \\
\midrule
\multirow{3}{*}{\shortstack[l]{Stepwise CoT\\vs.\ Prompting}} 
  & Overall    & 31 & ~3 & 26 & 54.2\% [41.7, 66.1] & 26.7\% \\
  & Soundness  & 29 & ~7 & 24 & 54.2\% [41.7, 66.1] & 31.7\% \\
  & Excitement & 30 & 10 & 20 & 58.3\% [45.7, 69.9] & 26.7\% \\
\bottomrule
\end{tabular}
}
\end{table}

\subsection{LLM Semantic Judge for FAS}
\label{sec:semantic_judge}
We use GPT-4.1-mini with a temperature of 0.1 to score the semantic similarity between a generated proposal and a candidate future paper. The score scale is 1-10. The full prompt is provided in Figure~\ref{fig:judge_prompt_fas}.

\begin{figure*}[t]
\centering
\begin{tcolorbox}[width=\textwidth,colback=gray!5,colframe=gray!60,boxrule=0.5pt,arc=2pt,left=4pt,right=4pt,top=4pt,bottom=4pt]
\footnotesize
\textbf{Prompt for Future-Alignment Scoring}

\medskip
You are an expert research paper evaluator. Your task is to score the semantic similarity between a generated research proposal and an existing paper, both overall and for specific subfields.

\medskip
\textbf{Generated Proposal:}\\
\{proposal\}

\medskip
\textbf{Candidate Paper:}\\
\{candidate\_summary\}

\medskip
\textbf{Task:}\\
Rate the similarity on a scale of 0--10 for each subfield and overall:
\begin{itemize}
    \item 0: Completely unrelated
    \item 3: Same broad area but different specific focus
    \item 5: Related with some overlapping ideas
    \item 7: Very similar direction
    \item 10: Nearly identical
\end{itemize}

Score each subfield:
\begin{enumerate}
    \item \textbf{research\_question}: How similar are the core research questions or problems being addressed?
    \item \textbf{hypothesis}: How similar are the hypotheses or expected outcomes?
    \item \textbf{proposed\_method}: How similar are the proposed methodologies or techniques?
    \item \textbf{novelty\_claims}: How similar are the claimed contributions and novelty?
    \item \textbf{experiment\_details}: How similar are the experimental setups, datasets, or evaluation approaches?
    \item \textbf{overall}: The holistic similarity considering all aspects
\end{enumerate}

\medskip
Respond with \textbf{only} a JSON object:
\begin{verbatim}
{
  "research_question": <0-10>,
  "hypothesis": <0-10>,
  "proposed_method": <0-10>,
  "novelty_claims": <0-10>,
  "experiment_details": <0-10>,
  "overall": <0-10>,
  "reasoning": "<brief explanation>"
}
\end{verbatim}
\end{tcolorbox}
\caption{Prompt used for LLM-based future-alignment scoring between a generated proposal and a candidate future paper.}
\label{fig:judge_prompt_fas}
\end{figure*}

\subsection{Multi-dimension LLM Judge}
\label{sec:ana_llm_judge}
We use GPT-4.1-mini with temperature 0 to evaluate proposal quality along three dimensions: resource validity, task--method consistency, and task--experiment consistency. The full prompt is provided in Figure~\ref{fig:judge_prompt_quality}.

\begin{figure*}[t]
\centering
\begin{tcolorbox}[width=\textwidth,colback=gray!5,colframe=gray!60,boxrule=0.5pt,arc=2pt,left=4pt,right=4pt,top=4pt,bottom=4pt]
\footnotesize
\textbf{Prompt for Proposal Quality Evaluation}

\medskip
You are a critical expert reviewer evaluating a research proposal. Be strict and skeptical. Score on 3 dimensions (1--5 scale).

\medskip
\textbf{Research Proposal:}\\
\{proposal\}

\medskip
\textbf{Evaluation Dimensions (be critical; average proposals should score 3):}

\begin{enumerate}
    \item \textbf{Resource Validity} (1--5): Are the mentioned datasets, benchmarks, baseline models, and tools real and correctly named?
    \begin{itemize}
        \item 5: All resources are verified as real, correctly named, and appropriate for the task
        \item 4: Most resources appear real, with at most minor naming issues or obscure references
        \item 3: A mix of clearly real resources and some that are generic, vague, or difficult to verify
        \item 2: Several resources appear fabricated, incorrectly named, or nonexistent
        \item 1: Most resources are clearly hallucinated or incorrectly described
    \end{itemize}
    \textbf{Red flags:} generic dataset names without specifics, made-up benchmark names, or nonexistent baseline models.

    \item \textbf{Task--Method Consistency} (1--5): Does the method actually solve the stated problem? Is there logical coherence?
    \begin{itemize}
        \item 5: The method directly addresses every aspect of the research question with a clear logical connection
        \item 4: The method addresses the main task but may have minor logical gaps
        \item 3: The method partially addresses the task, with some disconnected or tangential components
        \item 2: Weak connection between method and stated task, or the method solves a different problem
        \item 1: The method does not logically address the stated task
    \end{itemize}
    \textbf{Red flags:} method components that do not connect to the hypothesis, or solving a different problem than stated.

    \item \textbf{Task--Experiment Consistency} (1--5): Do the experiments actually validate the claims? Are the metrics appropriate?
    \begin{itemize}
        \item 5: The experiments are perfectly designed to test the hypothesis with appropriate metrics
        \item 4: The experiments are mostly appropriate, with only minor gaps in validation coverage
        \item 3: Some experiments do not match the task type, or some metrics are only partially appropriate
        \item 2: Significant mismatch; the experiments test different things than claimed
        \item 1: The experiments are completely inappropriate for the stated task
    \end{itemize}
    \textbf{Red flags:} wrong metrics for the task type, missing key experiments, or baselines from the wrong domain.
\end{enumerate}

\medskip
\textbf{Important:} Be critical. Most proposals have flaws. A score of 5 should be rare. Average proposals should score around 3.

\medskip
Respond with \textbf{only} a JSON object:
\begin{verbatim}
{
  "resource_validity": {"score": 1-5, "justification": "specific issues found or why it's good"},
  "task_method_consistency": {"score": 1-5, "justification": "specific logical gaps or strengths"},
  "task_experiment_consistency": {"score": 1-5, "justification": "specific mismatches or why appropriate"}
}
\end{verbatim}
\end{tcolorbox}
\caption{Prompt used for LLM-based proposal quality evaluation.}
\label{fig:judge_prompt_quality}
\end{figure*}


\end{document}